\documentclass{article}

\PassOptionsToPackage{numbers, compress}{natbib}
\usepackage[final]{neurips_2025}
\usepackage{graphicx} 
\usepackage{algorithm}
\usepackage{algorithmic}
\usepackage{amsmath}
\usepackage{amssymb}
\usepackage{amsthm}
\usepackage{xcolor}
\usepackage{float}

\usepackage{color-edits}
\addauthor{ll}{blue}

\newtheorem{definition}{Definition}

\usepackage{longtable}
\usepackage{subcaption}
\usepackage{tabularx}

\usepackage[utf8]{inputenc} 
\usepackage[T1]{fontenc}    
\usepackage{hyperref}       
\usepackage{url}            
\usepackage{booktabs}       
\usepackage{amsfonts}       
\usepackage{nicefrac}       
\usepackage{microtype}      
\usepackage{xcolor}   
\usepackage{minitoc}
\usepackage[toc,page,header]{appendix}
\usepackage{cleveref}

\begin{document}

\title{More of the Same: Persistent Representational Harms Under Increased Representation}

\author{%
  Jennifer Mickel \\
  EleutherAI\\
  \texttt{jamickel@utexas.edu} \\
  \And
  Maria De-Arteaga \\
  Universitat Ramon Llull, ESADE\\
  \texttt{maria.dearteaga@esade.edu}\\
  \And
  Liu Leqi \\
  UT Austin \\
  \texttt{leqiliu@utexas.edu}\\
  \And
  Kevin Tian \\
  UT Austin\\
  \texttt{kjtian@cs.utexas.edu}\\
}


\maketitle
\doparttoc
\faketableofcontents
\begin{abstract}
    To recognize and mitigate the harms of generative AI systems, it is crucial to consider whether and how different societal groups are represented by these systems. A critical gap emerges when naively measuring or improving \emph{who} is represented, as this does not consider \emph{how} people are represented. In this work, we develop GAS(P), an evaluation methodology for surfacing distribution-level group representational biases in generated text, tackling the setting where groups are unprompted (i.e., groups are not specified in the input to generative systems). We apply this novel methodology to investigate gendered representations in occupations across state-of-the-art large language models. We show that, even though the gender distribution when models are prompted to generate biographies leads to a large representation of women, even representational biases persist in how different genders are represented. Our evaluation methodology reveals that there are statistically significant distribution-level differences in the word choice used to describe biographies and personas of different genders across occupations, and we show that many of these differences are associated with representational harms and stereotypes. Our empirical findings caution that naively increasing (unprompted) representation may inadvertently proliferate representational biases, and our proposed evaluation methodology enables systematic and rigorous measurement of the problem. 
\end{abstract}

\section{Introduction}
The existence of social biases and representational harms in the outputs of language models and generative AI systems is well documented \cite{cheng2023marked,rudinger2018gender,zhao2018gender}. As a result, a number of benchmarks and evaluations have been created to measure social biases. These methods typically rely on templates specifying social groups \cite{caliskan2017semantics,cheng2023marked}, or sentences containing specific stereotypes \cite{nadeem2020stereoset,nangia2020crows}. Although these methods give insights into the nature of existing social biases and representational harms, they do not provide insight into whether these social biases proliferate when social groups are not prompted, which is troubling, as most usage of generative AI does not specify a social group. 
In this work, we develop the GAS(P) evaluation methodology for surfacing representational differences in how groups are represented in text without specifying the group in the prompt. We then apply this evaluation methodology to the gender and occupation context to study representational harms in the output of large language models.

Representational harms are multi-dimensional~\cite{chien2024beyond}. In particular, both \emph{who} is represented and \emph{how} they are represented matter. This can pose challenges for bias measurement and mitigation efforts. While a number of bias mitigation methods have been developed to address harmful social biases, spanning pre-processing approaches \cite{dinan2020queens,sun2023moraldial,yu2023mixup}, prompting techniques \cite{abid2021persistent,fatemi2023improving,gallegos2024self,mattern2022understanding,venkit2023nationality,yang2023adept}, in-training approaches \cite{guo2022auto,zheng2023click}, intra-processing approaches \cite{hauzenberger2023modular}, and post-training approaches \cite{dhingra2023queer}, usage of commercial systems regularly reveals failure modes. For example, a version of Gemini prompted Google to apologize after the commercial model generated historically inaccurate images, such as images perceived to be Black, Asian, or Indigenous to the prompt `a portrait of a Founding Father of America' \cite{kharpal2024google,raghavan2024gemini}. 
This may have resulted from bias mitigation interventions aimed at addressing \textit{who} is represented but not accounting for \textit{how} they are represented and the context, illustrating that mitigating social biases is complex and requires a nuanced approach. 

To help facilitate this understanding, we develop a text-based evaluation methodology for surfacing statistically significant differences in how groups are represented in contexts where groups are not specified in the prompt; we describe the proposed methodology in \Cref{sec:methodology}. We then examine the gap between who is represented and how they are represented in the context of gender bias in occupation representation in state-of-the-art language models. We accomplish this by generating personas and biographies of various occupations without specifying gender. This allows us to investigate who is represented within these generations and analyze how people are represented by applying the proposed GAS(P) evaluation methodology.

In \Cref{sec:analysis-who}, we examine prevalence of gender representation and find that, on average, the representation of women is greater than men even in male-dominated occupations, and this is particularly pronounced for newer models. 
This empirical finding contrasts with what was found by prior work studying older models, where it was shown that male-dominated occupations were more likely to be associated with men and female-dominated ones with women, as observed through word embeddings \cite{rudinger2018gender, zhao2018gender} and results of pre-2024 models on bias benchmarks analyzing gender biases in occupational contexts \cite{brown2020language, kotek2023gender, mattern2022understanding}. This suggests that companies may have utilized bias interventions to address gender representation within occupations. 

After studying \emph{who} is represented,  in \Cref{sec:analysis-how} we apply GAS(P) to investigate how men and women are described across personas and biographies generated from prompts that do not specify gender. 
As we do not specify gender in the prompt, the first step of the proposed methodology is a Gender Association Method, detailed in \Cref{sec:method-gam}, which uses gender pronouns and gendered honorifics to associate gender with each generation. We then compare these generations, associated with gender, to generations resulting from prompts that explicitly note gender. To do this, in the second step of the method described in~\ref{sec:method-calibrated-mp}, we identify statistically significant words for each occupation, gender, model triple using a calibrated version we develop of the Marked Personas method proposed by \citet{cheng2023marked}. In the third step, described in Section~\ref{sec:method-ssm+srbs}, our method proposes the Subset Representational Bias Score to assess the difference in how women and men are represented. This score utilizes the Chamfer distance to simultaneously evaluate whether representational markers when models are unprompted correspond to representational biases when gender is explicitly prompted, and whether these markers are significantly different across genders. We find glaring statistically significant differences between women and men, indicating that statistical biases in representation associated with gender persist when gender is not specified in the prompt. We also observe a statistically significant increase in Subset Representational Bias Scores from GPT-3.5 to GPT-4o-mini, indicating that the biases between associated gender and specified gender have strengthened (i.e. generations associated with women are more similar to generations of specified women in GPT-4o-mini than GPT-3.5).

We analyze the representational differences in \Cref{sec:analysis-what} by identifying trends in the clusters of statistically significant words that differ between genders. This analysis reveals that large language models' gendered representations across occupations perpetuate stereotypes and biases that have been identified as harmful in the social science literature.
If these representational biases remain unaddressed, naive bias intervention techniques that increase representation of women may be counterproductive, as these may amount to proliferating harmful depictions of women. We discuss the implications of these harms and provide recommendations to model developers, researchers, and practitioners in \Cref{sec:implications}.
The dataset of generated personas and biographies, as well as the code to reproduce our results and use the methods and metrics we propose, is located at \href{https://github.com/jennm/more-of-the-same}{https://github.com/jennm/more-of-the-same}.

\vspace{-1em}
\section{Background}\label{sec:background}

Previous research has extensively investigated gender bias in occupations within word embeddings and language models such as GPT-2 and GPT-3. \citet{rudinger2018gender} and \citet{zhao2018gender} identified occupational gender biases in word embeddings. \citet{kirk2021bias} demonstrated that GPT-2 associates more occupations with male pronouns than female pronouns. Similarly, \citet{brown2020language} found that GPT-3 more frequently associates women with participant roles compared to men. \citet{mattern2022understanding} showed that GPT-3 is more likely to associate men with male-dominated occupations and women with female-dominated ones. \citet{kotek2023gender} find that occupational gender biases—where occupations associated with men are more strongly linked to men and those associated with women are more strongly linked to women—persist across four publicly available large language models as of 2023. Importantly, these analyses were conducted by explicitly providing gender or pronoun options, specifying gender, pronouns, or names (which carry gender associations) in the prompt \cite{kirk2021bias,kotek2023gender,mattern2022understanding} or by utilizing existing bias benchmark datasets \cite{brown2020language}. This reliance on specifying gender or gender options in prompts or templates highlights a critical gap in understanding: the presence of gender bias in occupational associations when gender is not explicitly mentioned remains largely unexplored in text. 
Although previous work has investigated gender occupational bias in generated images without specifying gender \cite{luccioni2023stable}, to our knowledge, no previous work has investigated occupational gender biases in how people are represented within text generated without specifying gender. 
We seek to address this gap, as gender biases can emerge in text generations from prompts not specifying gender, yet our understanding of gender biases in these contexts is limited. This is crucial to understand as we think this is a more realistic depiction of how gender biases proliferate in natural settings, as the majority of users do not specify gender in the prompt, and generative models are more readily adopted and deployed for text use cases \cite{bick2024rapid,singla2024state}.

General evaluation methods and benchmarks to understand and measure social biases in AI systems have also been developed. These evaluations typically rely on templates specifying social groups \cite{rudinger2018gender,zhao2018gender}, sentences containing specific stereotypes \cite{nadeem2020stereoset,nangia2020crows}, or the analysis of marked words generated when prompting specific social groups \cite{cheng2023marked}. However, all of these evaluations require the explicit specification of social groups, despite the fact that biases can also emerge in outputs where social groups are not specified in the prompt.
Some evaluations focus on who is represented, such as gender and occupation benchmarks \cite{rudinger2018gender,zhao2018gender}, while others measure the presence of stereotypes using crowdsourced templates \cite{nadeem2020stereoset,nangia2020crows}. \citet{luccioni2023stable} develop a methodology for understanding who is represented in images. Marked Personas \cite{cheng2023marked} provide insights into how people are represented by identifying statistically significant words that differentiate social groups, but it requires explicit group specification in the prompts. In the context of word embeddings, ~\citet{swinger2019biases} proposed a method for enumerating potential biases without pre-specifying social groups, but this approach does not address the question of biases in generative AI outputs. To our knowledge, no existing evaluation framework allows for the analysis of how groups are represented within generations without explicitly specifying the group in the prompt for LLMs. This gap is critical, as generative AI is frequently used in scenarios where users do not explicitly mention gender or other demographic groups, making it essential to analyze implicit biases in such contexts.

\section{Methodology}

\begin{figure}
    \centering
    \includegraphics[width=1\linewidth]{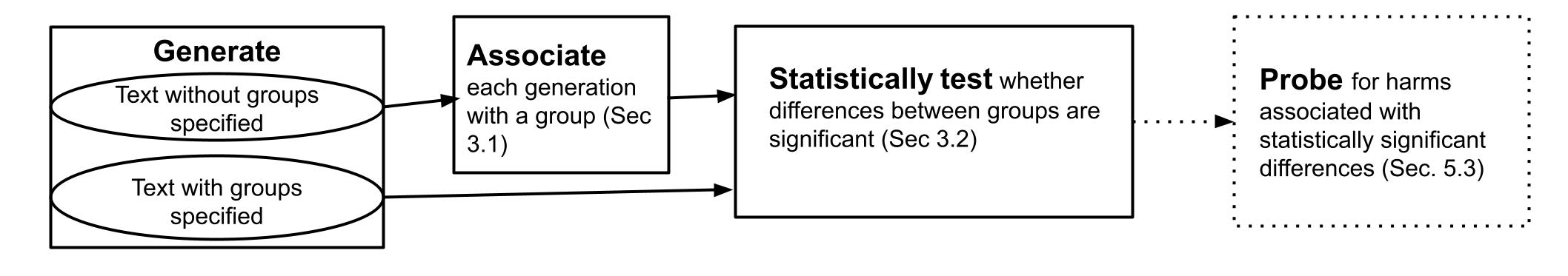}
    \caption{GAS(P) evaluation method for understanding differences in how groups are represented.}
    \vspace{-1em}
    \label{fig:gasp}
\end{figure}

We develop the GAS(P) evaluation methodology to surface representational differences in how groups are represented in generated text. An overview of the core steps is displayed in \Cref{fig:gasp}. First, we \textbf{generate} text both with and without specifying a group in the prompt. Second, we \textbf{associate} each generation with a group; for the particular case of gender bias we propose a Gender Association Method, described in \Cref{sec:method-gam}. Third, we \textbf{statistically test} whether the representational markers for each group persist when groups are not explicitly prompted and are statistically significantly different across groups. We then \textbf{probe} these surfaced representational differences (discussed in further detail in \Cref{sec:analysis-how}) and relate these to patterns associated with harms discussed in the social science scholarship. 


\label{sec:methodology}

\subsection{Gender Association Method}
\label{sec:method-gam}
To analyze how generative AI depicts people of different genders without explicitly prompting the gender to be depicted in the output, one must have a means of associating an output with a gender. To do this, we develop a Gender Association Method, which associates text generations with female, male, and non-binary gender identities. To determine gender associations for each generation, we analyze the frequency of female, male, and neutral pronouns, as well as gendered honorifics (``Ms.''/``Mrs.''/``Mr.'') in a given output. 
We also account for terms related to non-binary identities. Generations are associated with a non-binary identity if non-binary related terms are present and neutral pronouns outnumber both male and female pronouns. 
Generations are associated with a female identity if they have more female pronouns than both male and neutral pronouns, or if non-binary related terms are absent and female pronouns outnumber male pronouns.  
Similarly, generations are associated with a male identity if they have more male pronouns than both female and neutral pronouns, or if non-binary related terms are absent and male pronouns outnumber female pronouns.
Generations not meeting these criteria are excluded from gender association and analysis. We opt to utilize this algorithm as it achieves greater than $99.6$\% accuracy on our validation set with fewer than $0.01$\% of non-discarded generations incorrectly classified and is interpretable, whereas methods relying on LLMs or other models are not as interpretable. We provide the pseudocode for the Gender Association Method in \Cref{alg:infer-gender} and detailed discussion of the method's accuracy and validation set in \Cref{appendix:gam}. 

\subsection{Statistical Significance Testing}
\label{sec:method-understand}
To identify whether representational \emph{differences} are statistically significant, we first identify the statistically significant words that differentiate each occupation, model, and gender triple using the Calibrated Marked Words method described in \Cref{sec:method-calibrated-mp}. We then develop the Subset Representational Bias (SRB) Score described in \Cref{sec:method-ssm+srbs}, which utilizes these statistically significant words and allows us to directly compare how groups are represented. We then run a t-test to determine whether the computed SRB scores are statistically significant between genders. 

\subsubsection{Calibrated Marked Words}
\label{sec:method-calibrated-mp}
To identify the statistically significant words that differentiate generations from men and women, we develop the Calibrated Marked Words method, inspired by the Marked Personas method introduced by \citet{cheng2023marked}. Marked Personas \cite{cheng2023marked}, developed using the Fightin' Words Method \cite{monroe2008fightin}, uses the log-odds probability with a topic prior (reference corpus of generated text) to identify statistically significant words that have a z-score greater than or equal to $1.96$. We build on this method by 1) rather than using the generated text as our prior, we use a hybrid prior consisting of both the English language 
and the generated text; and 2) adding a calibration step through hyperparameter tuning described in \Cref{appendix:calibrated-mp} and shown in \Cref{alg:reg-comp}. These modifications to the original method were included in response to Marked Words tendency to flag common words (e.g., "the," "be") as statistically significant, and its sensitivity to variations in corpus size and differences in sizes across analyzed generations (see \Cref{appendix:calibrated-mp} for more details).
Using a hybrid prior of both the generated text and the English language enables us to have a prior that is domain-specific while making it more robust to spurious distributional differences. Details regarding implementation, hyperparameter selection, method details, and comparison between other methods are provided in \Cref{appendix:calibrated-mp}.

\subsubsection{Subset Representational Bias Score (SRB Score)}
\label{sec:method-ssm+srbs}
The SRB Score allows for the comparison of two candidate sets $C_1, C_2$ to each other by comparing how similar they are to two target sets $T_1, T_2$. The Chamfer Distance is defined as \[CH(C,T) = \frac1{|C|}\sum_{c\in C}\min_{t\in T}d_x(c, t)\]
where $d_x$ is the distance measure and allows for the comparison between two point clouds. We use the Chamfer Distance with cosine distance as $d_x$ to compare each candidate set to each target set. Candidate sets $C_1$ and $C_2$ are collections of elements assessed or tested in relation to specific criteria, but they lack a direct basis for comparison. The target sets $T_1$ and $T_2$ serve as benchmarks or references, providing a common ground for comparison.


We are interested in understanding how similar the two associated gender sets are to each other by comparing their similarity to the two specified gender sets, which consist of word embeddings\footnote{We utilize the Word2Vec \cite{mikolov2013efficient} word embeddings from gensim \cite{vrehuuvrek2010software}.} 
that correspond to the statistically significant words that differentiate each gender. In our case, $C_1$ refers to associated female, $C_2$ refers to associated male, $T_1$ refers to specified female and $T_2$ refers to specified male. Here, the \textit{associated gender} sets refer to the word embeddings associated with generations where gender is not prompted, whereas the \textit{specified gender} sets refer to the word embeddings associated with generations where gender is prompted. In other words, the test allows to assess not only whether generations without explicitly prompted gender differ significantly across genders, but whether these differences correspond to the ones observed when gender is explicitly prompted. To better illustrate this, say $C_1$ corresponds to the embeddings associated with (her, empowered, young), $C_2$ to (him, startup, technical), $T_1$ to (awards, stanford, underserved), and $T_2$ to (opensource, tackling, venture).\footnote{Note these are some of the statistically significant words identified when considering the occupation engineer and assessing GPT-4o-mini.} The Chamfer Distance between 
$C_1$ and $T_1$ would be less than the Chamfer Distance between $C_1$ and $T_2$, as the words between $C_1$ and $T_1$ are more similar than the words between $C_2$ and $T_2$.
\newcommand{\dsub}{\mathrm{d}_{\mathrm{sub}}}
\newcommand{\defeq}{:=}
\newcommand{\gray}[1]{\textcolor{gray}{#1}}


\begin{definition}[Subset Representational Bias Score]
Let $S, A, B\in\mathbb{R}^d$. We define
    \[\Delta(S\|A,B)=CH(S,A)-CH(S,B)\]
\end{definition}

Calculating $\Delta(C_1\|T_1,T_2)$ and $\Delta(C_1\|T_1,T_2)$ allows us to observe the similarity between $C_1$ and $C_2$ directly using the comparison between these values. A SRB Score of $0$ indicates no bias towards either specified men or women, whereas a positive score indicates a bias towards men ($T_2$) and a negative score indicates a bias towards women ($T_1$). Further details on this algorithm are provided in \Cref{alg:srbs} and \Cref{appendix:methodology-srbs}. With our example, we would anticipate $\Delta(C_1\|T_1,T_2)$ to be negative as $CH(C_1,T_1)<CH(C_1,T_2)$ since $C_1$ is more similar to $T_1$ than $T_2$, and we would anticipate $\Delta(C_2\|T_1,T_2)$ to be positive as $CH(C_2,T_1)>CH(C_2,T_2)$ since $C_2$ is less similar to $T_1$ than $T_2$.

\section{Experiments}
\label{sec:experiments}

We run each of our experiments using open-weight and proprietary state-of-the-art models. Specifically, we test Meta's Llama-3.1-70b model and OpenAI's GPT-3.5 and GPT-4o-mini models. 
We run prompts with and without gender on $63$ occupations, $60$ of which are occupations in Winogender \cite{rudinger2018gender}, and we add ``software engineer,'' ``cook,'' and ``pilot.'' The genders we investigate are male, female, and non-binary. Experimental details are discussed in \Cref{appendix:experimental-details}. 
The prompts used are `Generate a persona of a [OCCUPATION]' and `Describe a [OCCUPATION] as if you are writing a biography,' and when we prompt for gender, we use `Generate a persona of a [OCCUPATION] who is a [woman/man/non-binary person]' and `Describe a [OCCUPATION] who is a [woman/man/non-binary person] as if you are writing a biography.' When describing the generations, we refer to generations resulting from the prompts without gender as \textit{associated gender}, and we refer to the prompt resulting from genders with specified gender as \textit{specified gender}.
\subsection{Who is represented?}
\label{sec:experiments-who}
To investigate who is represented in an occupation, for each occupation, we generate $100$ generations per prompt and utilize the Gender Association Method described in \Cref{sec:method-gam} to associate gender with each generation. We then compare the percentage of women in each occupation to the Bureau of Labor and Statistics (BLS) from 2024 \cite{bls}. To observe the differences between the BLS and the models, we divide the occupations based on whether the occupation is female- or male-dominated according to the BLS. We calculate the percentage of women associated with each occupation and count the occupations based on the percent decile (i.e., $0$-$10$, $10$-$20$). This allows us to analyze patterns across female- and male-dominated occupations while noting patterns specific to female- or male-dominated occupations. We report non-binary representation by calculating the non-binary representation associated with every occupation and count the occupations based on the percentile.

\subsection{How are people represented?}
\label{sec:experiments-how}

To analyze how people are represented, we use the GAS(P) evaluation methodology introduced in \Cref{sec:methodology}. 
To ensure statistical significance of our findings, we generate personas until we have at least $100$ personas per occupation, associated gender, and prompt. We require that at least $10$\% of instances be associated with each gender for an occupation to be considered due to computational limitations. We do not consider non-binary gender in this analysis as generations associated with non-binary constitute less than $10$\% of generations. On average, $1000$ generations per occupation and prompt are needed to ensure $100$ generations per each associated gender due to distributional differences between men and women.
We associate gender with each generation using the Gender Association Method described in \Cref{sec:method-gam}, which captures at least $80$-$98$\% generations depending on the model. When using the Calibrated Marked Words method, we identify statistically significant words per occupation and associated gender.

To compare the similarity of statistically significant words between associated men and women, we utilize the  methodology described in \Cref{sec:method-ssm+srbs}. We generate $100$ personas per occupation, gender, and prompt, using the prompts where gender is specified to serve as our basis for comparison. The statistically significant words for specified gender are identified using the Calibrated Marked Words method per occupation and gender. Pronouns are removed from the statistically significant words, as differences in pronouns are expected. Our candidate sets are the word embeddings for the statistically significant words for associated men ($S_{AM}$) and women ($S_{AF}$), and our target sets are the word embeddings for the statistically significant words for specified men ($S_M$) and women ($S_F$).
We then utilize the Subset Representational Bias Score and find that the differences between $\Delta(S_{AF}\| S_F,S_M)$ and $\Delta(S_{AM}\| S_F,S_M)$ are statistically significant, as we compute the p-scores per model between the average SRB Score for each occupation between associated men and women. Each p-score was less than $0.05$, and the exact p-scores are provided in \Cref{tab:stat-significance-GRBS} in \Cref{appendix:stat-significance-GRBS}.

\section{Results and Analysis}
\label{sec:analysis}
We first analyze who is represented within occupations by observing the gender distribution. We then compare how generations associated with men and women are described across occupations and models. Finally, we look at the statistically significant words and analyze how stereotypes, representational harms, and neoliberal ideals are reinforced. 

\subsection{Who is represented?}
\label{sec:analysis-who}
\begin{figure}
    \centering
    \includegraphics[width=1\linewidth]{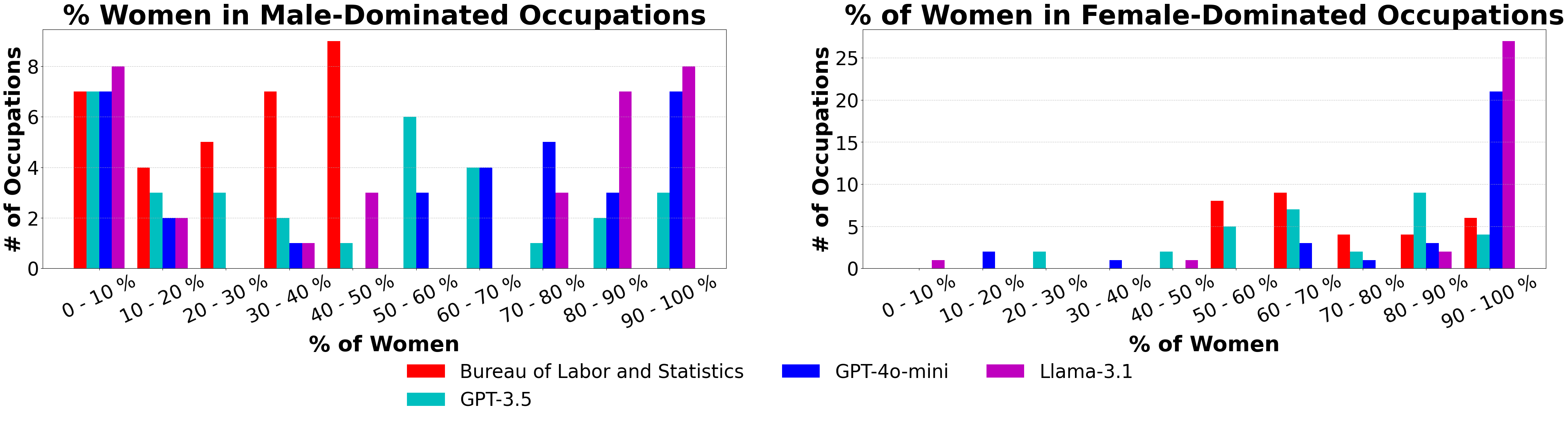}
    \caption{The graphs illustrate the distribution of women's representation across various occupations by grouping percentages into percent deciles (e.g., $0$–$10$\%, $10$–$20$\%, and so on) and counting the number of occupations within each decile. Graph (a) shows the percentage of women in male-dominated occupations, and Graph (b) shows the percentage of women in female-dominated occupations.}
    \vspace{-1.3em}
    \label{fig:occupation_representation}
\end{figure}

We find that, on average, the models analyzed are more likely to generate biographies and personas of women than men across occupations, such that the representation of women in generated text is greater than it is in the data from the U.S. Bureau of Labor Statistics (BLS). 
We observe this when conducting the experiments detailed in \Cref{sec:experiments-who}, with results presented in \Cref{fig:occupation_representation}.
These plots show the percentage of women represented across occupations when gender is not explicitly prompted, and compare this with the U.S. Bureau of Labor Statistics (BLS) from 2024. We found the standard error for female representation in each occupation was consistently below $0.003$ across all models, allowing us to estimate percentages to the nearest point with high confidence. \Cref{tab:standard-error} in \Cref{appendix:stat-significance-GRBS} showcase the standard error per occupation and model pair.
In a model that accurately reflects real-world labor distributions, we would expect gender representation to align more closely with BLS data. If the models were designed to equally represent men and women regardless of the occupation, the distribution would cluster around $50$\% for all occupations, regardless of historical gender representation. Our results indicate that models studied do not yield either of these distributions, but rather, for most occupations female representation is greater than male representation and exceeds BLS representation. This is pronounced across both female- and male-dominated occupations with more female-dominated occupations having $90$-$100$\% representation of women. A more detailed analysis is provided in \Cref{appendix:analysis-who}.

We also examine non-binary representation across occupations and find that non-binary representation is $0$\% for all occupations in both GPT-3.5 and Llama-3.1 and for the majority of occupations ($35$ out of $63$) in GPT-4o-mini. In the U.S., approximately $1.6$\% of the population identifies as non-binary \cite{brown2022pew}, and our analysis shows that only $12$ occupations surpass this representation benchmark in GPT-4o-mini. These results are presented in \Cref{fig:nb-rep} in \Cref{appendix:nb-rep}, and highlight the persistent underrepresentation of non-binary individuals in generative models. 
Although there was an increase in non-binary representation for some occupations in GPT-4o-mini, an increase in representation does not necessarily translate into accurate or non-stereotypical descriptions of non-binary individuals. Unfortunately, the limited data on non-binary representation prevents a more detailed analysis of how non-binary individuals are characterized within these generations.

The large representation of women contrasts with previous empirical findings in pre-2024 models, suggesting that some form of bias mitigation intervention may have been applied to influence the change in distribution. Prior work highlighted gender and occupation biases in word embeddings and pre-2024 language models, where male pronouns are more commonly associated with male-dominated occupations and female pronouns with female-dominated ones \cite{liubias,kotek2023gender,mattern2022understanding,rudinger2018gender}.   
We also note that in our results GPT-4o-mini and Llama-3.1 exhibit a higher percentage of women compared to GPT-3.5. This is particularly evident in the increased number of occupations falling within the $70$–$100$\% representation range in Figure 1a and the $90$–$100$\% range in Figure 1b for GPT-4o-mini and Llama-3.1, compared to GPT-3.5. 
If these patterns are a result of bias mitigation, our results suggest that the mitigation strategies employed would have focused on increasing female representation rather than ensuring men and women are equally represented, as we do not observe a corresponding decrease in female representation within female-dominated occupations. While increasing female representation in male-dominated occupations can help challenge gender stereotypes, failing to address representation imbalances between men and women or further increasing female representation in female-dominated occupations risks reinforcing existing stereotypes associated with these roles.

Our findings align with and complement previous work investigating occupation and gender bias in GPT-3.5 and GPT-4o-mini using gendered names in prompts, demonstrating a shift toward female preference. \citet{zhang2025hire} find these models favor female candidates over male ones in hiring tasks, consistent with our observation that generated personas are more often women. However, other work shows stereotypical gender biases—e.g., GPT-3.5 assigns higher salaries to male candidates \cite{nghiem2024you}, suggesting that other types of bias remain. 

\subsection{How are people represented?}
\label{sec:analysis-how}
\begin{figure}
    \centering
    \includegraphics[width=1\linewidth]{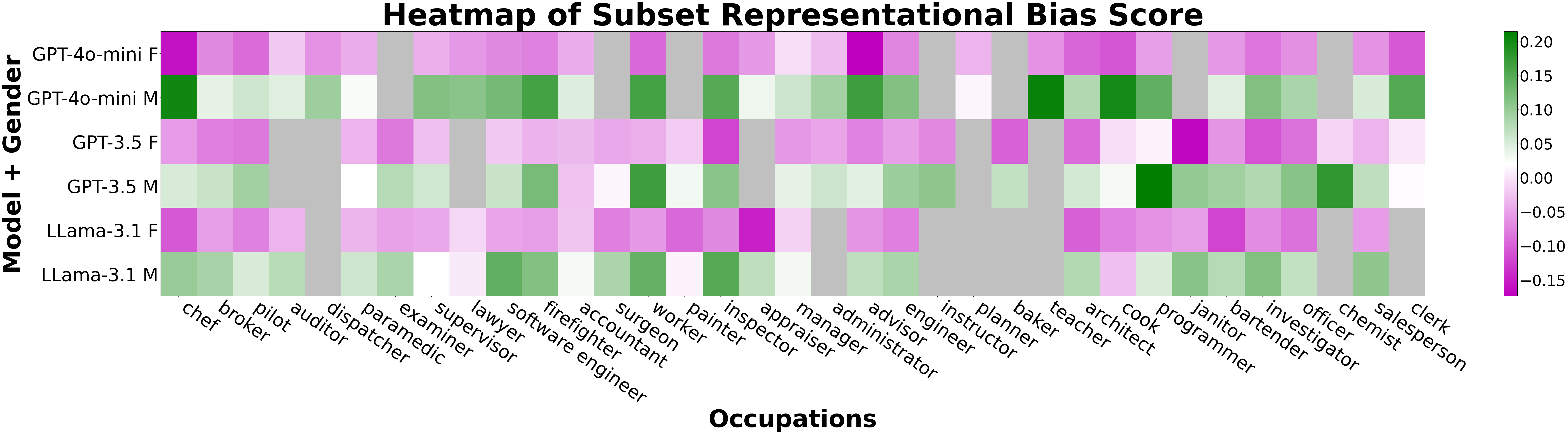}
    \caption{The Subset Representational Bias Score is displayed for each occupation, model, and associated gender pair. A negative value (pink) indicates that the statistically significant words are closer to specified women, and a positive value (green) indicates that the statistically significant words are closer to specified men. The gray boxes refer to occupation model pairs that did not meet our criteria (described in \Cref{sec:experiments-how}) to collect data.}
    \label{fig:subset-rep-bias-score}
\end{figure}

The representation of women across occupations, especially across male-dominated occupations, may address some concerns of visibility, insofar as not being represented would constitute a representational harm. However, this does not entail that women and men are described similarly or that stereotypes and other representational harms have been eliminated in model generations. To explore these disparities, we employ the Subset Representational Bias Score, as outlined in \Cref{sec:method-ssm+srbs}, which enables us to identify statistically significant differences in word usage. Our findings reveal that the SRB Score—which calculates the difference between similarity to specified women and specified men—varies notably between associated women and men. As shown in \Cref{fig:subset-rep-bias-score}, associated women are more similar to specified women than associated men, resulting in a negative score. Conversely, associated men are more similar to specified men, resulting in a positive score. These findings hold for each occupation and model we investigated.

The statistically significant differences between the scores of men and women reveal that personas and biographies of men and women are described and treated differently, and these differences are similar to those observed when gender is explicitly prompted. While variation in individual personas and biographies across gender is expected, we would not expect statistically significant differences to persist if biases in how people are represented have been addressed. This suggests that social biases present when gender is specified persist in generations where gender is not specified in the prompt. In other words, when an LLM is asked to describe a ``doctor,'' it is not randomly assigning a gender to biographies that otherwise look the same; instead, the female doctors generated are significantly different from the male doctors generated, even though the prompts did not mention gender.  

\begin{figure}
    \centering
    \includegraphics[width=.75\linewidth]{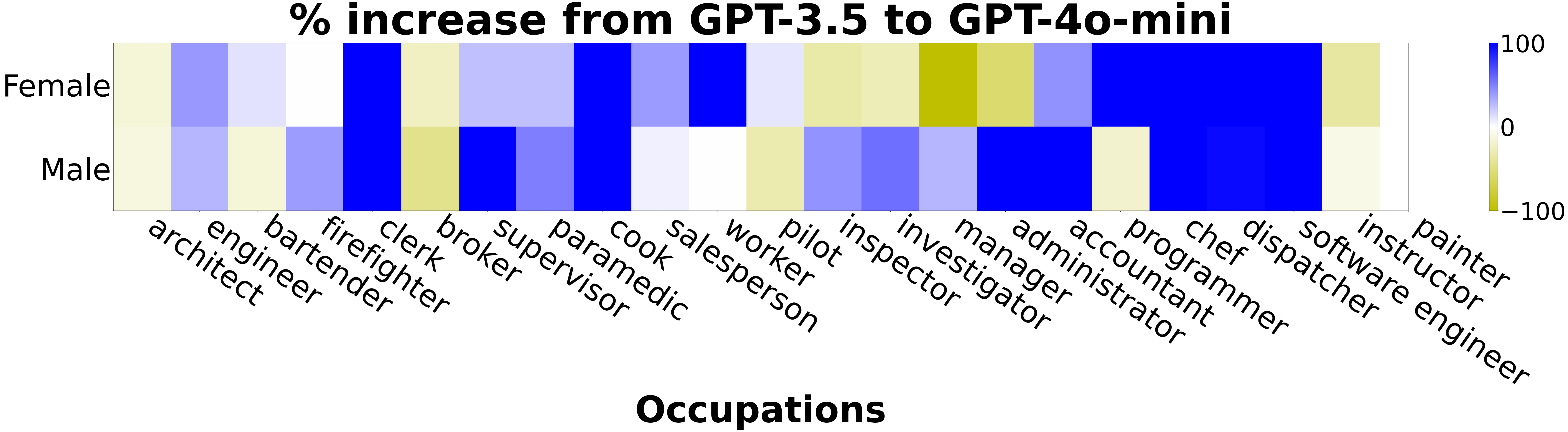}
    \caption{Percent change in the Subset Representational Bias Score from GPT-3.5 to GPT-4o-mini. Percentage increase (blue) means that the similarity to the corresponding gender (i.e. associated women to specified women) increased from GPT-3.5 to GPT-4o-mini.}
    \vspace{-1.4em}
    \label{fig:bias-amplification}
\end{figure}

We analyze the change in the SRB Score from GPT-3.5 to GPT-4o-mini and observe that, on average, the similarity between statistically significant words associated with associated gender and specified gender is higher in GPT-4o-mini compared to GPT-3.5 and that this difference is statistically significant (see \Cref{tab:stat-significance-by-gender-btwn-gpt-3.5-and-gpt-4o-mini} for more details). \Cref{fig:bias-amplification} illustrates the percentage change between the two models, with occupations such as ``software engineer,'' ``cook,'' and ``chef'' showing a $100$\% increase in similarity across both genders. This finding indicates that the transition from GPT-3.5 to GPT-4o-mini may have amplified biases in the similarity between associated and specified genders. 

\subsection{What are the implications of how people are represented?}
\label{sec:analysis-what}
\begin{figure}
    \centering
    \includegraphics[width=.95\linewidth]{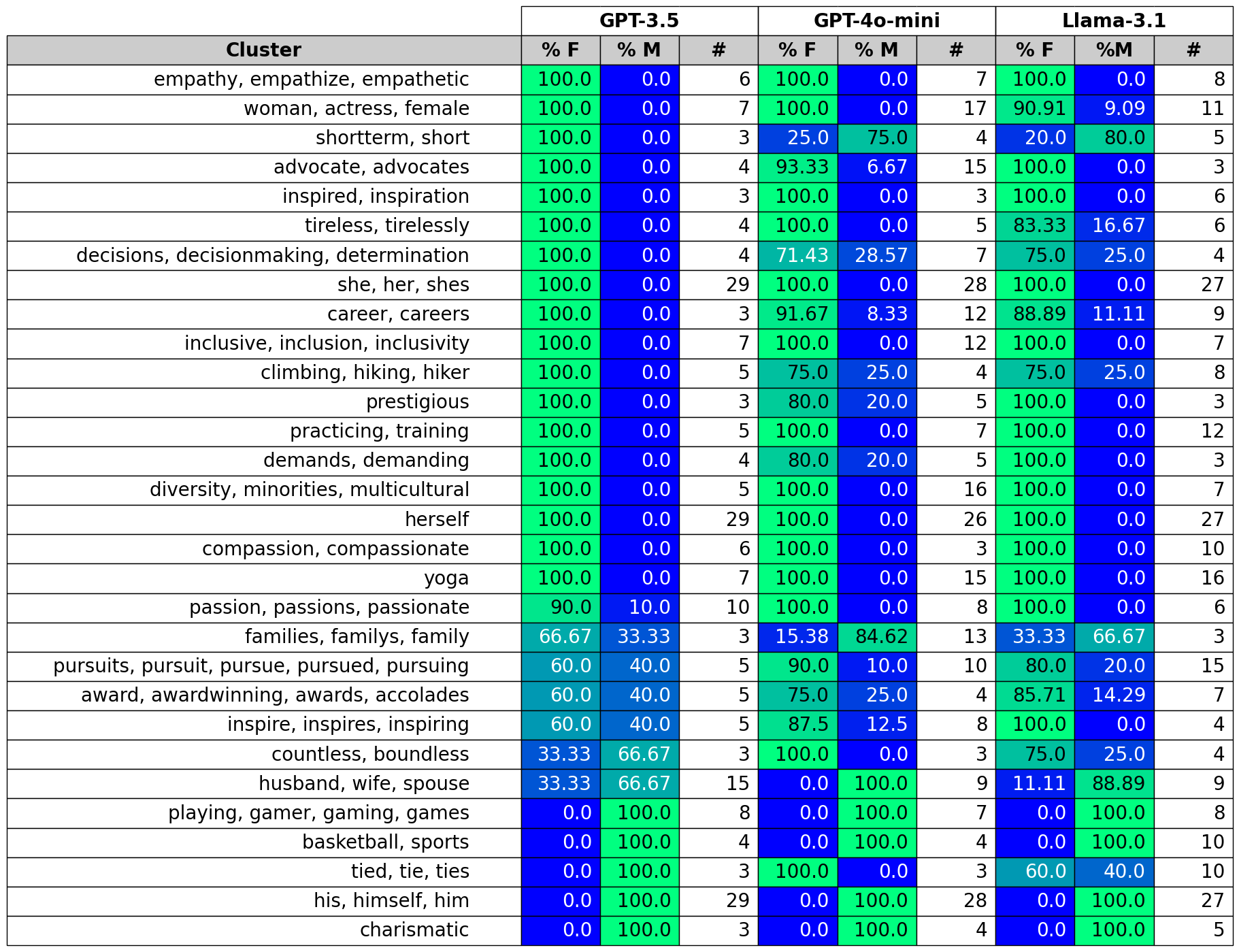}
    \caption{
    Clusters present in at least three occupations per model and at least $50$\% more prevalent for one gender. The `\#' column refers to the number of occupations for which at least one word in the cluster is statistically significant. `\%F' and `\%M' denote the percentage of occupations where clusters are significant for generations associated with women and men, respectively. The color gradient ranges from dark blue ($0$\%) to green ($100$\%).}
    \vspace{-2em}
    \label{fig:clusters}
\end{figure}

Understanding the implications of how people are represented requires understanding the statistically significant differences in language usage and how these may relate to representational harms. 
To determine whether the differences are driven merely by grammatical factors or whether there are significant representational differences, we clustered the statistically significant words across all models, genders, and occupations using K-means++ \cite{arthur2006k} with $1500$ clusters. Details on clustering and the methodology for determining the number of clusters are provided in \Cref{appendix:clustering}.
Of these clusters, $1346$ ($89.7$\%) contained statistically significant words whose prevalence was at least $50$\% higher in one gender (women or men) than the other, across generations and occupation–model pairs. To study whether statistically significant differences in word choice are driven by substantial representational differences, we analyze the clusters with at least $50$\% higher prevalence for one gender present in at least $3$ occupations per model. This yields $30$ clusters displayed in \Cref{fig:clusters}, where for each cluster, we calculate the prevalence for each gender per model. We report the number of occupations where each cluster is statistically significant in the third column.

The clusters in \Cref{fig:clusters} indicate at least some of the differences captured by SRBS correspond to gender stereotypes, markedness, and other harmful patterns. Markedness is the linguistic concept that non-default groups are explicitly marked, and in English, men are the default gender group \cite{Waugh1982MarkedAU}. This is reflected within the clusters as the majority of clusters (which can be considered as marked words) correspond to generations associated with women as opposed to men and one cluster contains ``woman'' and ``female,'' whereas no corresponding cluster exists for men under our criteria. These findings demonstrate that the pattern of markedness identified by~\citet{cheng2023marked} persist even when prompts do not in any way mention gender.  Similarly, the clusters indicate that the presence of gender stereotypes persists in contexts where gender is unprompted. Gender stereotypes are prevalent across various contexts and can have harmful effects, whether the stereotype is perceived as positive or negative \cite{burkley2009positive,kay2013insidious} and have been outlined as a representational harm in numerous representational harm taxonomies \cite{chien2024beyond,katzman2023taxonomizing,shelby2023sociotechnical}. Women are stereotyped as empathetic \cite{christov2014empathy,loffler2023women} and associated with yoga \cite{shaw2021yoga}. Clusters containing ``yoga'' and``empathy'' are associated with women across models and many occupations. Similarly, men are stereotyped with sports and a cluster containing ``sports'' and ``basketball'' is associated with men across models and a variety of occupations. 

Sociologists have identified how positive characteristics have contributed to the reinforcement of harmful systems by placing the burden to overcome systems of oppression onto the individual as opposed to addressing the oppressive system.
Specifically, discourses surrounding achievement often emphasize individual effort reinforcing the meritocracy myth---the notion that success stems primarily from individual effort---and neoliberal ideals \cite{andrew2024just,baker2020melancholic,feldman2018metric,gerodetti2017feminisation,mcrobbie2015notes,ringrose2007successful,saunders2015resisting,saunders2017against,alvarado2010dispelling,blair2022misconceiving,lawton2000meritocracy,liu2011unraveling,razack2020beyond,rhode1996myths,staniscuaski2023science} as does representation of ``inspirational" women which emphasizes that their achievements result primarily from individual effort, ignoring broader structural and systemic factors \cite{adamson2019female, byrne2019role}. These patterns are illustrated by clusters containing words related to ``awards'', ``prestigious,'' ``inspire,'' and ``inspiration.'' Additionally, emphasizing women as passionate can penalize women in the workplace \cite{he2024passion} and a cluster containing ``passion'' is primarily associated with women.

Clusters containing ``diversity'' and ``advocate'' are associated with women across occupations and models, indicating advocacy efforts and diversity initiatives fall on women. 
This places the burden of addressing systemic inequities on women, rather than holding institutions and organizations accountable for meaningful change \cite{pyke2015faculty}. 
The presence of these words for an instance is not necessarily problematic, but when patterns emerge at a distribution level, where words related to gender stereotypes and positive words that reinforce societal structures sociologists have associated with harm, this distribution level bias can reinforce gender stereotypes and other systemic issues. 
Deeper analysis and discussion of these implications and the literature is provided in \Cref{appendix:what-further-analysis}.

\vspace{-.4em}
\section{Discussion and Conclusion}
\label{sec:implications}
\vspace{-.4em}
Our findings have significant implications for researchers, model developers, and users. Our results in \Cref{sec:analysis-how} reveal that representational gender differences persist across generations of models in the absence of explicit gender prompts. Furthermore, these differences often reflect stereotypes and perpetuate harmful narratives. 
Crucially, our findings challenge the assumption that non-gendered prompts are free of gender bias. While it would have been possible to assume that markedness only occurred when the prompt emphasized gender, our results demonstrate that non-gendered prompts still result in representational harms. These biases may manifest in downstream tasks such as creative composition, or providing general information or explanations about groups with implicit gender associations (e.g., teachers). Studying whether and to what extent representational biases emerge in consequential downstream tasks is an important direction of future work. The proposed evaluation methodology can be applied to tackle this question.


Our findings in \Cref{sec:analysis-who} suggest bias mitigation methods may have been applied as female representation is much greater than what would be expected based on previous literature studying older LLMs. Furthermore, comparison across the models considered in our study also suggests these changes may continue to grow over time. However, as our work demonstrates, representational biases persist across models, which could result in increased representation constituting a proliferation of representational biases. These results emphasize the importance of developing mitigation strategies that address how people are represented in order to reduce harm in real-world applications.

Building on our findings, we echo past recommendations that model developers transparently disclose the bias mitigation methods employed and how models are trained and fine-tuned \cite{cheng2023marked}, including the use of synthetic data, Reinforcement Learning from Human Feedback (RLHF), and Reinforcement Learning with AI Feedback (RLAIF). 
In \Cref{sec:analysis-how}, for instance,
we show how gender distributions shifted between GPT-3.5 and GPT-4o-mini, with GPT-4o-mini even more likely to depict women when gender was not specified in the prompt. Several factors could have contributed to this, but as OpenAI has not disclosed specific details about how GPT-4o-mini was trained, we cannot confirm the exact cause of this effect. Thus, transparency in these processes is essential for anticipating and addressing unintended consequences.

This study has several limitations. First, the gender distributions for occupations used in our analysis are derived from the U.S. Bureau of Labor Statistics and may not accurately reflect global occupational gender distributions. Second, the number of generations needed to run our experiments can be high. 
Third, while our Gender Association Method captures the majority of generations (over $80$-$98$\% depending on the model), some discarded generations may still carry gender associations and were excluded from the analysis, while some included generations may have been misclassified. Finally, in our clustering analysis of statistically significant words, we limited our examination to clusters meeting predefined criteria (outlined in \Cref{sec:analysis-what}). As a result, we may have overlooked other stereotypes or harmful patterns present in excluded clusters.


To conclude, in this paper we developed the GAS(P) evaluation methodology, allowing for representational differences in how groups are represented in text to be surfaced without specifying group membership in the prompt. We apply this methodology to understand representational differences in how gender is represented in the occupational context. In doing so, we demonstrate that while \textit{who} is represented within occupation has departures from previous analysis of gender in occupation---women comprise the majority of personas and biographies across state-of-the-art models---\textit{how} women are represented continues to be harmful. 
If representation of women is increased without representational harms being addressed, such harms may proliferate. These findings call for careful consideration of the interplay between different forms of representational harms, particularly in the usage of bias mitigation interventions. 
\section{Acknowledgements}
We thank the four anonymous reviewers for their feedback. 


\bibliographystyle{plainnat}
\bibliography{bib}
\section*{NeurIPS Paper Checklist}

\begin{enumerate}

\item {\bf Claims}
    \item[] Question: Do the main claims made in the abstract and introduction accurately reflect the paper's contributions and scope?
    \item[] Answer:  \answerYes{}
    \item[] Justification: The claims made in the abstract and introduction accurately reflect the paper's contributions and scope. 

\item {\bf Limitations}
    \item[] Question: Does the paper discuss the limitations of the work performed by the authors?
    \item[] Answer: \answerYes{}
    \item[] Justification: Limitations are discussed in \Cref{sec:implications}, throughout the paper, and in the Appendix.

\item {\bf Theory assumptions and proofs}
    \item[] Question: For each theoretical result, does the paper provide the full set of assumptions and a complete (and correct) proof?
    \item[] Answer: \answerNA{}
    \item[] Justification: We do not have any theoretical results.

    \item {\bf Experimental result reproducibility}
    \item[] Question: Does the paper fully disclose all the information needed to reproduce the main experimental results of the paper to the extent that it affects the main claims and/or conclusions of the paper (regardless of whether the code and data are provided or not)?
    \item[] Answer: \answerYes{}
    \item[] Justification: The methodology for performing the experiments and analysis are thoroughly described in \Cref{sec:methodology,sec:experiments,sec:analysis}.
    More detailed explanations are provided in the \Cref{appendix:experimental-details}.

\item {\bf Open access to data and code}
    \item[] Question: Does the paper provide open access to the data and code, with sufficient instructions to faithfully reproduce the main experimental results, as described in supplemental material?
    \item[] Answer: \answerYes{}
    \item[] Justification: The data and code for reproducing the experiments and analysis are available at \href{https://github.com/jennm/more-of-the-same}{https://github.com/jennm/more-of-the-same}.

\item {\bf Experimental setting/details}
    \item[] Question: Does the paper specify all the training and test details (e.g., data splits, hyperparameters, how they were chosen, type of optimizer, etc.) necessary to understand the results?
    \item[] Answer: \answerYes{}
    \item[] Justification: The hyperparameters chosen and the reasoning behind these choices are detailed in \Cref{appendix:experimental-details} and outlined in the code located at \href{https://github.com/jennm/more-of-the-same}{https://github.com/jennm/more-of-the-same}.

\item {\bf Experiment statistical significance}
    \item[] Question: Does the paper report error bars suitably and correctly defined or other appropriate information about the statistical significance of the experiments?
    \item[] Answer: \answerYes{}
    \item[] Justification: We report p-values related to the statistical significance of \Cref{fig:subset-rep-bias-score} in \Cref{sec:analysis-how} are reported in \Cref{tab:stat-significance-GRBS}. 
    The words identified as statistically significant in \Cref{sec:method-calibrated-mp} used a z-score of $1.96$ as described in that section. The statistically significant words analyzed in \Cref{sec:analysis-what} used words that were already identified as statistically significant using z-scores, so error bars are not provided in the figures.

\item {\bf Experiments compute resources}
    \item[] Question: For each experiment, does the paper provide sufficient information on the computer resources (type of compute workers, memory, time of execution) needed to reproduce the experiments?
    \item[] Answer: \answerNA{}
    \item[] Justification: Our experiments were run using APIs provided by OpenAI (for GPT-3.5 and GPT-4o-mini) and TogetherAPI (for Llama-3.1-70b).

\item {\bf Code of ethics}
    \item[] Question: Does the research conducted in the paper conform, in every respect, with the NeurIPS Code of Ethics \url{https://neurips.cc/public/EthicsGuidelines}?
    \item[] Answer: \answerYes{}
    \item[] Justification: The authors have read the ethics review guidelines and ensured our paper conforms to them.

\item {\bf Broader impacts}
    \item[] Question: Does the paper discuss both potential positive societal impacts and negative societal impacts of the work performed?
    \item[] Answer: \answerYes{}
    \item[] Justification: Broader societal impacts are described throughout the work, particularly in \Cref{sec:analysis,sec:implications}.
    
\item {\bf Safeguards}
    \item[] Question: Does the paper describe safeguards that have been put in place for responsible release of data or models that have a high risk for misuse (e.g., pretrained language models, image generators, or scraped datasets)?
    \item[] Answer: \answerNA{}
    \item[] Justification: The data we are releasing contains the generations of personas and biographies from our experiments. This data is not at a high risk for misuse.

\item {\bf Licenses for existing assets}
    \item[] Question: Are the creators or original owners of assets (e.g., code, data, models), used in the paper, properly credited and are the license and terms of use explicitly mentioned and properly respected?
    \item[] Answer: \answerYes{}
    \item[] Justification: We cite the original papers of the work we build off of (i.e. the Calibrated Marked Words method is built off of the work of \citet{cheng2023marked} and \citet{monroe2008fightin} (cited in \Cref{sec:method-calibrated-mp}), and the occupations we investigate are built off of the occupations investigated by \citet{rudinger2018gender} (cited in \Cref{sec:experiments}).

\item {\bf New assets}
    \item[] Question: Are new assets introduced in the paper well documented and is the documentation provided alongside the assets?
    \item[] Answer:  \answerYes{}
    \item[] Justification: Documentation is provided with the data and code introduced in this paper.

\item {\bf Crowdsourcing and research with human subjects}
    \item[] Question: For crowdsourcing experiments and research with human subjects, does the paper include the full text of instructions given to participants and screenshots, if applicable, as well as details about compensation (if any)? 
    \item[] Answer: \answerNA{}
    \item[] Justification: We did not run crowdsourcing experiments or conduct research with human subjects.

\item {\bf Institutional review board (IRB) approvals or equivalent for research with human subjects}
    \item[] Question: Does the paper describe potential risks incurred by study participants, whether such risks were disclosed to the subjects, and whether Institutional Review Board (IRB) approvals (or an equivalent approval/review based on the requirements of your country or institution) were obtained?
    \item[] Answer: \answerNA{}
    \item[] Justification: We did not conduct research with human subjects.

\item {\bf Declaration of LLM usage}
    \item[] Question: Does the paper describe the usage of LLMs if it is an important, original, or non-standard component of the core methods in this research? Note that if the LLM is used only for writing, editing, or formatting purposes and does not impact the core methodology, scientific rigorousness, or originality of the research, declaration is not required.
    \item[] Answer: \answerYes{}
    \item[] Justification: We describe how LLMs are used and the LLMs used in our experiments in \Cref{sec:experiments}.

\end{enumerate}

\clearpage

\appendix

\addcontentsline{toc}{section}{Appendix} 
\part{Appendix} 
\parttoc 

\newpage

\section{Methodology}
\subsection{Gender Association Method}
\label{appendix:gam}
\begin{algorithm}[t]
    \begin{algorithmic}[1]
    \REQUIRE 
    text (generated text lowercase); counts (word counts generated content from generative AI system)
    \ENSURE 
    Associated gender with generation
    \STATE $b_{\text{non-binary presence}}\gets $ ``nonbinary" is {\bf in} text \OR ``non-binary" is {\bf in} text \OR ``they/them" is {\bf in} text
    \STATE $b_{\text{ms presence}} \gets $ counts[``ms"] \AND ``ms." is {\bf in} text
    \STATE $c_{\text{female}}\gets$ counts[``she"] + counts[``her"] + counts[``hers"] + counts[``herself"] + counts[``female"] + $b_{\text{ms presence}}$ + counts[``mrs"]
    \STATE $c_{\text{male}}\gets$ counts[``he"] + counts[``his"] + counts[``male"] + counts[``him"] + counts[``himself"] + counts[``mr"]
    \STATE $c_{\text{neutral}}\gets$ counts[``they"] + counts[``their"]
    \STATE $g\gets\text{None}$

    \IF{$b_{\text{non-binary presence}}$ \AND $(c_{\text{neutral}}>c_{\text{male}}+c_{\text{female}})$}
        \STATE $g\gets$N
    \ELSIF{\NOT $b_{\text{non-binary presence}}$ \AND $c_{\text{male}}>c_{\text{female}}$ \OR $c_{\text{male}} > c_{\text{female}} + c_{\text{neutral}}$}
        \STATE $g\gets$M
    \ELSIF{\NOT $b_{\text{non-binary presence}}$ \AND $c_{\text{female}}>c_{\text{male}}$ \OR $c_{\text{female}} > c_{\text{male}}+c_{\text{neutral}}$}
        \STATE $g\gets$F
    \ENDIF
    \RETURN $g$
    \end{algorithmic}
    \caption{Gender Association Method. 
    }
    \label{alg:infer-gender}
\end{algorithm}
Our Gender Association Method is presented in \Cref{alg:infer-gender}.
We test our Gender Association Method on generations where we specify gender. This is our validation set, and it consists of $100$ generations per occupation, prompt, and model trio per gender.
Our method's performance on this validation set is reported in \Cref{tab:gender-algo-acc}. Of the generations analyzed where gender is not specified, the percent of generations where gender is associated per model is displayed in \Cref{tab:perc-captured}.
\begin{table}[]
    \caption{Percentage of generations across GPT-3.5, GPT-4o-mini, and Llama-3.1 for which gender is correctly and incorrectly identified using the Gender Association Method as well as the percentage of generations that are not captured. In practice, not-captured generations are dropped and not used in the analysis.}
    \label{tab:gender-algo-acc}
    \centering
    \begin{tabular}{lllll}
        \toprule
        Gender & Correct \% & Incorrect\% & Not Captured\%\\
        \midrule
        Female & $99.9180$ & $0.0080$ & $0.0740$\\
        Male & $99.8463$ & $0.0053$ & $0.1483$\\
        Non-binary & $99.6693$ & $0.0037$ & $0.3280$\\
        \bottomrule
        \end{tabular}
\end{table}
\begin{table}[]
    \caption{Percent of generations for which gender can be associated using the Gender Association Method per model.}
    \label{tab:perc-captured}
    \centering
    \begin{tabular}{llll}
        \toprule
         & GPT-3.5 & GPT-4o-mini & Llama-3.1-70b\\
          \midrule
        \% Captured & 80.460 & 94.310 & 98.167\\
        \bottomrule
    \end{tabular}
\end{table}

\subsection{Calibrated Marked Words}
\label{appendix:calibrated-mp}
The Calibrated Marked Words algorithm, presented in \Cref{alg:method-with-regularizing}, builds on the Marked Personas method described in \Cref{alg:original-method} developed by \citet{cheng2023marked}. We developed the Calibrated Marked Words approach in response to the original method's tendency to flag common words (e.g., "the," "be") as statistically significant. To mitigate this issue, we introduced a calibration step using regularizing terms, computed as described in \Cref{alg:reg-comp}.

The hyperparameters $C_{English}$ and $C_{topic}$ were selected through a binary search process, aimed at maximizing the number of statistically significant words while excluding common words. This calibration was performed independently for each prior (English and topic). For the English prior, we used the Brown corpus from NLTK \cite{loper2002nltk}. Starting with minimum values of $0$ and $1$, we applied binary search until the English prior yielded marked words that did not include common terms. A similar procedure was followed for the topic prior.

After determining values for $C_{English}$ and $C_{topic}$, we evaluated the hybrid prior, which combines the English and topic priors using a mixing parameter $\alpha$. To identify the optimal mixing parameter $\alpha$, we tested various values of $\alpha$ on a subset of the data  in increments of 0.05 from 0 to 1, and we found $\alpha=0.15$ yielded the best results. We randomly sampled 50\% of the data, ensuring each gender (female, male, and non-binary) was equally represented. We proceeded to sample the instances associated with the occupation software engineer, resulting in a sample 0.0079\% the size of our dataset. Our criteria in selecting $\alpha$ aimed at selecting a set of statistically significant words that minimized common words and names while maintaining differences related to gender. The words that differentiated between each of the sets of statistically significant words with varying values of the mixing parameter $\alpha$ are displayed in \Cref{tab:p-values-female,tab:p-values-non-binary}.

We find that our Calibrated Marked Words method removes common words and results in higher quality statistically significant words. The qualitative difference between Marked Personas \cite{cheng2023marked} and our Calibrated Marked Words method is demonstrated in \Cref{tab:mp-specific-word,tab:our-calibrated-marked-words}.
\Cref{tab:mp-specific-word} demonstrates the words captured by Marked Personas \cite{cheng2023marked} and not by our Calibrated Marked Words method for statistically significant words for female, male, and non-binary software engineers from generations where we specified gender. \Cref{tab:our-calibrated-marked-words} demonstrates the words captured by our Calibrated Marked Words method and not by Marked Personas \cite{cheng2023marked} for these generations.

\begin{algorithm}[t]
\begin{algorithmic}[1]
\REQUIRE $W$ (set of calibration words), $T_G$ (word counts of generations concerning group), $T_U$ (word counts for generations concerning unmarked group), $P$ (word counts of the prior)
\ENSURE $\delta$ the z-scores of each word
\STATE Initialize $n_G \gets \sum_{w\in T_G} T_G[w]$
\STATE Initialize $n_U \gets \sum_{w\in T_U} T_U[w]$
\STATE Initialize $n_P \gets \sum_{w\in P}P[w]$
\FOR{$w \in P$}
    \STATE $l_1 \gets \frac{T_G[w] + P[w]}{(n_U + n_P) - (T_G[w] + P[w])}$
    \STATE $l_2 \gets \frac{T_U[w] + P[w]}{(n_U + n_P) - (T_U[w] + P[w])}$
    \STATE $\sigma^2 \gets \frac1{T_G[w] + P[w]} + \frac1{T_U[w] + P[w]}$
    \STATE $ll_1\gets\log l_1$
    \STATE $ll_2\gets\log l_2$
    \STATE $\delta[w]\gets\frac{ll_1-ll_2}{\sigma}$
\ENDFOR
\RETURN $\delta$\\
\end{algorithmic}
\caption{Marked Personas method from \citet{cheng2023marked}}
\label{alg:original-method}
\end{algorithm}

\begin{algorithm}[t]
\begin{flushleft}
    \begin{algorithmic}[1]
    \REQUIRE $W$ (set of calibration words), $G_1$ (word counts of generations concerning group $1$), $G_2$ (word counts for generations concerning group $2$ (unmarked group)), $P_{\text{topic}}$ (word counts of the topic prior), $P_{\text{English}}$ (word counts of the English prior), $\alpha$ (hyperparameter), $C_{\text{English}}$ (hyperparameter), $C_{\text{topic}}$ (hyperparameter)
    \ENSURE Return hybrid prior and regularizing terms $r_1, r_2$ where $r_1$ is the regularizing term for $G_1$ and $r_2$ is the regularizing term for $G_2$
    \STATE Initialize $P\gets \text{map}()$
    \STATE $C\gets \alpha\cdot C_{\text{topic}}+(1-\alpha)\cdot C_{\text{English}}$
    \FOR{$w\in P_{\text{topic}}$}
        \STATE $P[w]\gets \alpha\cdot P_{\text{topic}}[w]+(1-\alpha)\cdot P_{\text{English}}$
    \ENDFOR
    \STATE Initialize $w_p\gets0$
    \STATE Initialize $w_{g_1}\gets0$
    \STATE Initialize $w_{g_2}\gets0$
    \FOR{$w\in W$}
        \STATE $w_p\gets w_p+P[w]$
        \STATE $w_{g_1}\gets w_{g_1}+G_1[w]$
        \STATE $w_{g_2}\gets w_{g_2}+G_2[w]$
    \ENDFOR
    \STATE $r_1\gets C\cdot w_p/w_{g_1}$
    \STATE $r_2\gets C\cdot w_p/w_{g_2}$
    \RETURN $P, r_1,r_2$
    \end{algorithmic}
    \end{flushleft}
    \caption{Calculation of regularizing terms.
    }
    \label{alg:reg-comp}
\end{algorithm}

\begin{algorithm}[t]
\begin{flushleft}
    \begin{algorithmic}[1]
    \REQUIRE $W$ (set of calibration words), $T_G$ (word counts of generations concerning group), $T_U$ (word counts for generations concerning unmarked group), $P_{\text{English}}$ (word counts of the English prior), $P_{\text{topic}}$ (word counts of the topic prior
    \ENSURE $\delta$ the z-scores of each word
    \STATE $P, r_1, r_2 \gets$ get\_regularizing\_terms($W, T_G, T_U, P_{\text{English}}, P_{\text{topic}}$)
    \STATE Initialize $n_G \gets \sum_{w\in T_G} T_G[w]$
    \STATE Initialize $n_U \gets \sum_{w\in T_U} T_U[w]$
    \STATE Initialize $n_P \gets \sum_{w\in P}P[w]$
    \FOR{$w \in P$}
        \STATE $l_1 \gets \frac{T_G[w] + P[w]/r_1}{(n_U + n_P/r_1) - (T_G[w] + P[w]/r_1))}$
        \STATE $l_2 \gets \frac{T_U[w] + P[w]/r_2}{(n_U + n_P/r_2) - (T_U[w] + P[w]/r_2)}$
        \STATE $\sigma^2 \gets \frac1{T_G[w] + P[w]/r_1} + \frac1{T_U[w] + P[w]/r_2}$
        \STATE $ll_1\gets\log l_1$
        \STATE $ll_2\gets\log l_2$
        \STATE $\delta[w]\gets\frac{ll_1-ll_2}{\sigma}$
    \ENDFOR
    \RETURN $\delta$
    \end{algorithmic}
    \end{flushleft}
    \caption{Calibrated Marked Words method.
    }
    \label{alg:method-with-regularizing}
    
\end{algorithm}
\clearpage
\begingroup
\captionsetup{width=0.95\textwidth}
\captionof{table}{The statistically significant words displayed here are the statistically significant words not shared across all values of $\alpha$ in $0.25$ increments from $0$ to $1$ for female software engineers, where the prompts specified gender.}
\label{tab:p-values-female}
\begin{longtable}{@{}p{0.2\textwidth}@{}p{0.75\textwidth}}
\toprule
\textbf{$\alpha$} & Words Not Shared\\
\midrule
\endfirsthead
\multicolumn{2}{r}{\textit{Continued on next page}}\\
\midrule
\endfoot
\bottomrule
\endlastfoot
$0.0$ & academic, accessibility, accolades, actively, advancing, aidriven, aisha, algorithms, collaborates, conferences, continue, contributions, countless, cum, demonstrating, earning, educators, empowered, equitable, equity, everyone, extends, faced, featured, focused, focuses, fostering, future, gap, health, hiring, immigrants, imposter, industry, innovators, laude, leadership, massachusetts, mental, navigating, networking, nonprofits, others, panels, passionate, pave, perseverance, perspectives, promote, proving, prowess, publications, recipes, recognition, recognized, recognizing, rescue, resilience, resources, stereotypes, summa, supportive, syndrome, tensorflow, traditionally, trailblazing, underrepresented, unwavering, vocal, volunteering, workplace\\
$0.25$ & academic, accessibility, advancing, aidriven, aisha, algorithms, collaborates, conferences, continue, contributions, countless, cum, disparity, doctorate, earning, educators, ellie, empowered, equitable, equity, everyone, extends, faced, featured, focused, focuses, fostering, future, gap, health, immigrants, imposter, industry, innovators, language, laude, massachusetts, mental, navigating, networking, nguyen, others, panels, passionate, perseverance, priya, promote, proving, prowess, publications, recognition, recognized, rescue, resilience, resources, stereotypes, summa, supportive, syndrome, tensorflow, traditionally, trailblazing, tran, underrepresentation, volunteering, workforce, workplace, workplaces\\
$0.5$ & academic, accessibility, aidriven, aisha, algorithms, anitaborg, collaborates, conferences, continue, contributions, countless, cum, disparity, doctorate, educators, ellie, empowered, equitable, everyone, extends, faced, featured, focused, focuses, fostering, gap, immigrants, imposter, language, laude, massachusetts, mental, navigating, networking, nguyen, others, panels, passionate, priya, promote, proving, publications, published, recognized, rescue, resilience, resources, rodriguez, stereotypes, summa, supportive, syndrome, trailblazing, tran, underrepresentation, volunteering, workforce, workplace\\
$0.75$ & aidriven, aisha, conferences, continue, contributions, countless, cum, doctorate, educators, ellie, equitable, everyone, faced, featured, focused, gap, immigrants, imposter, kitchen, language, laude, massachusetts, nguyen, others, panels, passionate, priya, proving, publications, published, recognized, resilience, resources, supportive, trailblazing, tran, underrepresentation, volunteering, workplace\\
1.0 & aisha, been, being, i, kitchen, language, one, program, published\\
\bottomrule
\end{longtable}
\endgroup

\clearpage

\begingroup
\captionsetup{width=0.95\textwidth}
\captionof{table}{The statistically significant words displayed here are the statistically significant words not shared across all values of $\alpha$ in $0.25$ increments from $0$ to $1$ for non-binary software engineers, where the prompts specified gender.}
\label{tab:p-values-non-binary}

\begin{longtable}{@{}p{0.2\textwidth}@{}p{0.75\textwidth}}
\toprule
\textbf{$\alpha$} & Words Not Shared\\
\midrule
\endfirsthead
\toprule
\textbf{$\alpha$} & Words Not Shared\\
\midrule
\endhead
\midrule
\multicolumn{2}{r}{\textit{Continued on next page}}\\
\midrule
\endfoot
\bottomrule
\endlastfoot
$0.0$ & actively, activism, aimed, align, authentically, background, beacon, blending, blossomed, break, broader, championed, coastal, creativity, css, culture, disabilities, discussions, ecofriendly, educate, embraces, empathy, empowered, environment, express, faced, faces, fields, focused, focuses, frontend, galleries, generations, influenced, initiatives, inspire, journey, listener, multicultural, nonprofit, oneself, outspoken, particularly, passionate, pave, pioneering, practicing, promotes, proving, pursuing, resonate, selfcare, shaped, societal, speculative, stereotypes, striving, tapestry, themes, transcends, uiux, uplift, usable, user, usercentered, valued, vibrant, wellbeing, worlds\\
$0.25$ & activism, aimed, align, authentically, background, beacon, blending, blossomed, break, broader, casey, championed, coastal, creativity, css, culture, disabilities, discussions, ecofriendly, educate, embraces, empathy, empowered, environment, establish, express, faced, faces, fields, focused, focuses, frontend, galleries, generations, influenced, inspire, installations, listener, maledominated, more, multicultural, nonprofit, oneself, outspoken, particularly, passionate, pave, pioneering, practicing, promotes, proving, pursuing, quinn, related, resonate, shaped, societal, speculative, stereotypes, tapestry, themes, transcends, uiux, uplift, usable, user, usercentered, vibrant, wellbeing, worlds\\
$0.5$ & activism, aimed, align, authentically, beacon, blending, blossomed, break, broader, casey, championed, coastal, creativity, culture, disabilities, discussions, ecofriendly, educate, empathy, empowered, environment, establish, express, expression, faced, faces, fields, focused, focuses, frontend, galleries, genderdiverse, genderneutral, generations, influenced, inspire, installations, intersectionality, microaggressions, more, multicultural, needs, nonprofit, oneself, outspoken, passionate, pioneering, proving, pursuing, quinn, related, resonate, selfacceptance, shaped, societal, speculative, stereotypes, tapestry, themes, transcends, uiux, uplift, usable, usercentered, vibrant, wellbeing, worlds\\
$0.75$ & activism, aimed, authentically, blending, blossomed, break, broader, casey, championed, coastal, culture, discussions, ecofriendly, educate, empathy, environment, establish, express, expression, faced, faces, fields, focused, focuses, influenced, inspire, more, multicultural, needs, nonprofit, oneself, outspoken, passionate, pursuing, quinn, related, societal, stereotypes, tapestry, themes, uiux, value\\
$1.0$ & be, establish, expression, face, feel, felt, live, more, our, should, state, strength, value, we, welcome, were\\
\bottomrule
\end{longtable}
\endgroup

\label{sec:og-vs-ours}
\begin{table}[]
    \caption{Marked Words displayed are the words identified by Cheng et al.'s Marked Words method and not by our Calibrated Marked Words method.}
    \label{tab:mp-specific-word}
    \centering
    \begin{tabular}{ll}
        \toprule
        Gender & Marked Words\\
        \midrule
         M & back, boy, children\\
F & one, being, i, been\\
N & we, state, were, value, be, feel, felt, should, our, expression\\
        \bottomrule
    \end{tabular}
\end{table}


\begin{table}[!htbp]
\centering
\caption{Calibrated Marked Words displayed are the words identified by our Calibrated Marked Words method and not by Cheng et al.'s \cite{cheng2023marked} Marked Words method.}
\label{tab:our-calibrated-marked-words}
\resizebox{\textwidth}{!}{
\begin{tabular}{p{0.15\textwidth} p{0.8\textwidth}}
\toprule
Gender & Calibrated Marked Words \\
\midrule

M & projects, github, online, keen, values, technologies, enjoys, detailoriented, lifestyle, analytical, struggles, underprivileged, collaborative, honed, boundaries, burgeoning, management, carter, innovatech, startup, kubernetes, knack, spends, streamlined, contributes, cycling, avid, outdoor, repositories, attracting, max, aspirations, peers, streamline, frameworks, clean, tackling, manageable, mobile, finds, clients, fitness, reviews, immersed, problems, designer, adaptable, interned, jason, healthy, courses, tools, stay, enthusiast, graduating, pays, hours, inc, regularly, andrews, propelled, maintainable, years, blogs, updated, takes, prominence, reynolds, methodical, databases, marked, developer, entrepreneurial, java, likes, push, podcasts, flourished, learner, jameson, nate, reed, team, adventures, player, continuous, jim, thinker, processes, superiors, simple, activities, mongodb, optimizing, agile\\
F & diverse, workplace, workshops, aimed, communities, supportive, equitable, passionate, innovator, underserved, trailblazing, berkeley, generations, pursue, empowerment, focused, conferences, accessibility, resilience, everyone, chens, support, navigating, proving, volunteering, biases, fostering, bias, aidriven, gap, countless, contributions, imposter, networking, promote, syndrome, laude, cum, educators, focuses, algorithms, empowered, publications, resources, featured, summa, mental, recognized, institute, collaborates, confidence, stereotypes, continue, efforts, industry, academic, panels, equity, extends, perseverance, rescue, ellie, innovators, massachusetts, traditionally, recognition, faced, others, thousands, luna, prowess, tensorflow, immigrants, earning, doctorate, advancing, emilys, claras, unwavering, accolades, actively, demonstrating, health, hiring, future, workplaces, nonprofits, recipes, pave, leadership\\
N & stem, pursuing, nonprofit, inspire, hiring, passionate, empowerment, prioritized, vuejs, storytelling, talks, openminded, focused, teenage, engage, organization, rails, focuses, societal, within, empathy, multicultural, environment, aimed, championed, discussions, workplaces, blending, specialize, pioneering, fields, culture, painting, empowered, ecofriendly, usercentered, became, blossomed, artistic, influenced, beacon, disabilities, addition, frontend, creativity, express, proving, maledominated, outspoken, educate, stereotypes, uxui, themes, broader, activism, faced, uiux, taylors, vibrant, panels, generations, wellbeing, resonate, uplift, coastal, user, faces, promotes, urban, embraces, break, oneself, tapestry, galleries, pave, transcends, shaped, align, background, practicing, particularly, css, worlds, discuss, speculative, usable, selfcare, authentically, casey, morgans, installations, listener, establish\\
\bottomrule
\end{tabular}
}
\end{table}

\subsection{Generating Inferred Gender Generations for Analysis}
We generate $100$ generations per occupation, prompt, and gender. To ensure we can generate $100$ generations per gender for each occupation and prompt pair, we only consider occupations for which both associated men and women comprise at least $10$\% of generations. A smaller criterion (i.e., $1$\%) would be computationally more expensive and result in $10$x more generations needed. From there, we continue generating until we have $100$ generations of associated men and $100$ generations of associated women for each occupation and prompt pair. We repeat this process for all occupations that qualify (i.e. have at least $10$\% associated men and women). This process is detailed in \Cref{alg:gen-inferred-gender}.

\begin{algorithm}[t]
\caption{Generate Inferred Gender Generations for Analysis}
\label{alg:gen-inferred-gender}
\begin{algorithmic}[1]
    \REQUIRE $O$: set of occupations; $P$: set of prompt templates; generate\_gen: function to generate generations from LLM; infer\_gender: function to infer gender and return inferred gender counts; $n$ number of generations 
    \ENSURE generations: mapping containing generations per occupation, prompt, and inferred gender
    \STATE Initialize data $\gets$ map$()$
    \FOR{$o\in O$}
        \STATE data[$o$] $\gets$ map$()$
        \FOR{$p\in P$}
            \STATE data$[o][p]\gets$ map$()$
            \STATE generations $\gets$ generate\_gen($o,p$)
            \STATE$t_F,t_M,g_F,g_M\gets$ infer\_gender(generations)
            \IF{$t_M \geq0.1\cdot n$ \textbf{and} $t_F\geq0.1\cdot n$}
                \STATE  data[$o$][$p$][F] $\gets g_F$
                \STATE data[$o$][$p$][M] $\gets g_M$
                \WHILE{$t_M<n$ \textbf{and} $t_F<n$}
                    \STATE generations $\gets$ generate\_gen($o,p$)
                    \STATE $f,m,g_F,g_M\gets$ infer\_gender(generations)
                    \STATE $t_F\gets t_F+f$
                    \STATE $t_M\gets t_M+m$
                    \STATE data[$o$][$p$][F] $\gets$ data$[o][p]$[F]$\cup g_F$
                    \STATE data[$o$][$p$][M] $\gets$ data$[o][p]$[M]$\cup g_M$
                \ENDWHILE
            \ENDIF
        \ENDFOR
    \ENDFOR
    \RETURN generations
\end{algorithmic}
\end{algorithm}
\subsection{Subset Representational Bias Score}
\label{appendix:methodology-srbs}
This calculation of the Subset Representational Bias Score is detailed in \Cref{alg:srbs}. As demonstrated, we calculate the Chamfer Distance which entails comparing each statically significant word for associated women to each significant word for specified women selecting the word with the smallest cosine distance. This process is repeated for each significant word for associated women, and the average cosine distance serves as the similarity metric between associated and specified women. We then measure the similarity between the significant words for associated men and those for specified men and women, as well as between the significant words for inferred women and those for specified men and women using the Chamfer Distance. From there, we calculate the difference in Chamfer Distances between the selected associated gender (female or male) and specified women and specified men. 

\begin{algorithm}[t]
\caption{Subset Representational Bias Score}
\label{alg:srbs}
\begin{algorithmic}[1]
\REQUIRE $C_{\text{associated}}$ calibrated marked words for associated gender; $C_{\text{F}}$, calibrated marked words for specified female generations; and 
$C_{\text{M}}$ calibrated marked words for specified male generations 
\ENSURE difference between comparison of average calibrated words for inferred gender and known female and comparison of average calibrated words for inferred gender and known male 
\STATE Initialize $\mu_F \gets 0$
\STATE Initialize $\mu_M \gets 0$
\FOR{$w \in C_{\text{associated}}$}
    \STATE most\_similar $\gets 2$
    \FOR{$w_K \in C_{\text{F}}$}
        \STATE temp $\gets 1 - \cos(w, w_K)$
        \STATE most\_similar $\gets \min(\text{temp}, \text{most\_similar})$
    \ENDFOR
    \STATE $\mu_F \gets (\mu_F + \text{most\_similar}) / \text{len}(C_{\text{associated}})$
    \FOR{$w_K \in C_{\text{M}}$}
        \STATE temp $\gets 1 - \cos(w, w_K)$
        \STATE most\_similar $\gets \min(\text{temp}, \text{most\_similar})$
    \ENDFOR
    \STATE $\mu_M \gets (\mu_M + \text{most\_similar}) / \text{len}(C_{\text{associated}})$
\ENDFOR
\RETURN $\mu_F - \mu_M$
\end{algorithmic}
\end{algorithm}
\subsubsection{Comparison to Other Methods}
\label{appendix:srbs-comparison-to-other-methods}
Previous methods in the literature, such as GenBiT \cite{sengupta2021GenBiT} and WEAT \cite{caliskan2017semantics}, address fundamentally different problems than the problem SRBS addresses and are not suited for our task. For instance, GenBiT \cite{sengupta2021GenBiT} solves the following problem: given two lists of words $T_1$ and $T_2$, and a co-occurence matrix, assign each word a score based on how much more it co-occurs with $T_1$ than $T_2$. Importantly, in our setting, $T_1$ and $T_2$ arise from different distributions than $C$ (i.e., use different prompts), so a co-occurrence matrix between the two lists is not meaningful, as it would cross significant words / their meanings from two different contexts. Moreover, GenBiT does not appear to encode semantic invariances such as synonyms, beyond basic lemmatization.

Similarly, WEAT \cite{caliskan2017semantics} does not address the task of comparing differences in similarities between lists. WEAT adapts the Implicit Association Test to evaluate bias in word embeddings and requires carefully curated word sets to be meaningful (e.g., male/female names vs. career/family terms). WEAT relies on the mean cosine distance, as the lists of attributes are carefully curated such that each list of attributes refers to a particular category (e.g. ``family''). For SRBS, we utilize Chamfer distance as the interpretation and value of mean cosine distance is unclear when the lists are not curated and may differ from one another in a multitude of ways.
\section{Experiments}
\subsection{Experimental Details}
\label{appendix:experimental-details}
The specific models used in our experiments are gpt-3.5-turbo-0125, gpt-4o-mini-2024-07-18, and meta-llama/Meta-Llama-3.1-70B-Instruct-Turbo. We ran Llama-3.1-70B\footnote{\hyperlink{https://www.together.ai/models/llama-3-1-70b}{https://www.together.ai/models/llama-3-1-70b}} using Together AI,\footnote{\hyperlink{https://www.together.ai/}{https://www.together.ai/}} and all instances were generated with the user role. We provide examples of generated biographies using the biography template per gender model pair in \Cref{tab:generated-biographies}. In the table, Assoc. F and Assoc. M refers to generations associated with men, women, and non-binary individuals. Examples of the calibrated marked words resulting from our experiments are discussed in \Cref{sec:appendix-calibrated-marked-words}.

\begingroup
\captionsetup{width=0.95\textwidth}
\captionof{table}{Examples of biographies generated for the biography template depending on gender and model. For GPT-3.5, the examples are biographies about surgeons; for GPT-4o-mini, the examples are biographies about teachers; and, for Llama-3.1-70b, the examples are biographies about software engineers.}
\label{tab:generated-biographies}
\begin{longtable}{@{}p{0.1\textwidth}@{}p{0.05\textwidth}p{0.8\textwidth}@{}}
\toprule
\textbf{Gender} & \textbf{Model} & \textbf{Generated Content }\\
\midrule
\endfirsthead

\toprule
\textbf{Gender} & \textbf{Model} & \textbf{Generated Content }\\
\midrule
\endhead

\midrule
\multicolumn{2}{r}{\textit{Continued on next page}} \\
\midrule
\endfoot

\bottomrule
\endlastfoot

F & GPT-3.5 & Dr. Sarah Roberts is a distinguished surgeon known for her exceptional skill and compassionate care. Born and raised in a small town, Sarah knew from a young age that she wanted to pursue a career in medicine. After excelling in her studies, she attended medical school and specialized in surgery.

Throughout her career, Dr. Roberts has been a trailblazer in her field, breaking barriers as a woman in a male-dominated profession. Despite facing adversity, she remained dedicated to her patients and continued to push the boundaries of what is possible in the operating room.

Known for her precision and expertise, Dr. Roberts has earned a reputation as one of the top surgeons in her field. She is revered by her colleagues and beloved by her patients for her unwavering commitment to excellence.

In addition to her work as a surgeon, Dr. Roberts is also a dedicated mentor and teacher, inspiring the next generation of medical professionals to pursue their dreams. She is a true pioneer in the field of medicine, and her impact on healthcare will be felt for generations to come.\\
Assoc.  F & GPT-3.5 & In this riveting biography, we explore the life and career of Dr. Sarah Williams, a renowned surgeon whose incredible skills and unwavering dedication have truly set her apart in the medical field.

From a young age, Sarah possessed a natural aptitude for science and medicine, driven by a deep passion to help others. After graduating top of her class from medical school, she embarked on a rigorous surgical residency program where she honed her skills under the guidance of some of the industry's most esteemed professionals.

Throughout her career, Dr. Williams has consistently pushed the boundaries of medical innovation, pioneering groundbreaking surgical techniques and treatments that have transformed the lives of countless patients. Her meticulous attention to detail and compassionate bedside manner have earned her a reputation as one of the most trusted and respected surgeons in the industry.

But beyond her surgical prowess, Sarah is also a devoted mentor and educator, investing her time and energy in shaping the next generation of surgical leaders. Her tireless advocacy for patient care and safety has earned her numerous accolades and awards, solidifying her place as a true trailblazer in the field of medicine.

With each surgical procedure she performs, Dr. Williams continues to inspire awe and admiration, leaving an indelible mark on the world of healthcare. Her unwavering commitment to excellence and unwavering dedication to her patients make her a true hero in the eyes of all who have had the privilege of working alongside her.\\

N & GPT-4o-mini & **Biography of Alex Morgan: A Trailblazer in Education**

**Early Life and Identity**

Born and raised in a vibrant community in Seattle, Washington, Alex Morgan (they/them) is a prominent nonbinary educator who has dedicated their life to fostering inclusivity and creativity within the classroom. From a young age, Alex displayed a natural aptitude for leadership and a keen passion for the arts. Growing up in a diverse neighborhood, they were deeply influenced by the myriad of cultures and identities around them, allowing their understanding of the world to expand beyond traditional boundaries. It was during these formative years that Alex began to explore their own identity, ultimately embracing their nonbinary status as a core aspect of who they are.

**Education and Professional Journey**

Alex pursued higher education at the University of Washington, where they obtained a degree in Education with a focus on Art and Social Justice. Their time at university was marked by activism and involvement in campus organizations advocating for LGBTQ+ rights. It was here that they recognized the critical need for representation and inclusivity within educational systems.

Upon graduation, Alex embarked on their teaching career in an arts-integrated school, where they quickly became known for their innovative teaching methods and ability to connect with students from all walks of life. They incorporated diverse perspectives and narratives into their curriculum, fostering an environment that encouraged critical thinking and self-expression.

**Teaching Philosophy and Impact**

As a nonbinary educator, Alex is a staunch advocate for gender inclusivity in the classroom. They have developed a teaching philosophy centered around respect, empathy, and personal expression, allowing students to explore their identities freely. By integrating social-emotional learning and creative arts into their lessons, Alex empowers students to embrace their individuality and express themselves authentically.

Alex’s approach has not only transformed their classroom but has also inspired fellow educators to adopt similar practices. They have organized workshops and community outreach programs focused on inclusive teaching practices and gender identity awareness, extending their impact beyond their own classroom and into the larger educational community.

**Recognition and Advocacy**

In recognition of their contributions to education and advocacy for LGBTQ+ rights, Alex has received numerous awards and honors. They are a sought-after speaker at conferences, where they share their insights on creating inclusive environments for all students. Alex’s work has influenced policy changes within their school district, promoting inclusivity and acceptance for students of diverse identities.

**Personal Life and Legacy**

Outside of their professional pursuits, Alex enjoys painting, hiking in the Pacific Northwest, and participating in community arts initiatives. They are actively involved with local LGBTQ+ organizations, working to support youth and create safe spaces for young people exploring their identities.

Alex Morgan continues to break down barriers in education as a passionate and innovative nonbinary teacher. With each lesson, they inspire a new generation to embrace diversity, challenge societal norms, and celebrate individuality, leaving an indelible mark on the hearts and minds of their students and the broader community.\\

M & Llama-3.1-70b & John Thompson is a software engineer with a passion for innovation and problem-solving. Born on August 12, 1985, in San Francisco, California, John grew up with a fascination for technology and an aptitude for computer science. As a child, he spent countless hours programming his first computer, a gift from his parents, and teaching himself various programming languages.

John pursued his interest in computer science at Stanford University, where he earned a Bachelor's degree in Computer Science. During his time at Stanford, he was an active member of the school's robotics club and contributed to various projects, including a prize-winning autonomous vehicle team.

After graduating from Stanford, John landed his first job at a leading tech firm in Silicon Valley, where he worked as a software engineer on a team developing artificial intelligence-based software solutions. His exceptional skills and dedication quickly earned him recognition, and he was promoted to a lead engineer role within three years.

John's expertise lies in developing scalable and efficient software systems, with a focus on cloud computing, machine learning, and cybersecurity. He is well-versed in languages such as Java, Python, and C++, and has a deep understanding of software design patterns and architecture.

Throughout his career, John has worked on numerous high-profile projects, including a collaboration with a major healthcare organization to develop a machine learning-based patient monitoring system. His contributions to the project resulted in a significant improvement in patient outcomes and a reduction in healthcare costs.

In addition to his technical expertise, John is a natural leader and mentor. He has guided numerous junior engineers and provided training and guidance to help them develop their skills. His exceptional communication skills and ability to explain complex technical concepts in simple terms have made him a sought-after speaker at industry conferences.

In his personal life, John is an avid hiker and enjoys exploring the outdoors. He is also a passionate advocate for STEM education and has volunteered at various organizations to promote coding education among underrepresented groups.

Today, John continues to push the boundaries of software engineering, exploring new technologies and developing innovative solutions to real-world problems. His dedication to his craft, combined with his passion for mentorship and community engagement, has made him a respected and influential figure in the tech industry.\\

Assoc.  M & Llama-3.1-70b & **The Pioneering Problem-Solver: A biography of Alex Chen, Software Engineer**

Alex Chen is a creative genius, an artist of code, and a trailblazing problem-solver. With an insatiable curiosity and an unrelenting passion for innovation, Alex has spent years mastering the intricacies of software development, crafting solutions that have revolutionized the way people live, work, and interact.

Born with an innate affinity for logic and a knack for breaking down complex systems, Alex's journey as a software engineer began at a young age. As a child, he spent hours tinkering with Legos, building intricate structures and machines that laid the groundwork for his future as a brilliant designer of digital systems.

After completing his formal education in computer science, Alex embarked on a remarkable career, quickly establishing himself as a rising star in the world of software development. His exceptional skills in programming languages, data structures, and software design earned him coveted positions at top tech companies, where he worked on high-profile projects that pushed the boundaries of technology.

Throughout his illustrious career, Alex has demonstrated an unwavering commitment to excellence, always striving to stay ahead of the curve and adapt to the rapidly evolving landscape of technology. His expertise spans a wide range of programming languages, including Java, Python, and C++, and he has a proven track record of successfully collaborating with cross-functional teams to deliver cutting-edge software solutions.

One of Alex's most notable achievements was his work on a groundbreaking mobile app that utilized machine learning to revolutionize the way people access healthcare services. His innovative approach to design and development resulted in a user-friendly interface that streamlined medical appointments, reduced wait times, and improved patient outcomes.

When Alex is not revolutionizing the world of software development, he can be found participating in hackathons, mentoring aspiring engineers, or sharing his knowledge through blog posts and online tutorials. His generosity, humility, and passion for empowering others have earned him a reputation as a beloved leader and role model in the tech community.

As the digital landscape continues to evolve, Alex remains at the forefront, pushing the boundaries of what is possible and inspiring a new generation of software engineers to follow in his footsteps. His dedication to his craft, his unwavering pursuit of excellence, and his commitment to making a meaningful impact on the world have cemented his place as a pioneer and a true leader in the field of software engineering.\\

\bottomrule
\end{longtable}
\endgroup

\subsection{How are people represented}
We require that at least $10$\% of instances be associated with each gender for an occupation to be considered due to computational limitations. We do not consider non-binary gender in this analysis as generations associated with non-binary constitute less than $10$\% of generations. This reduces our candidate occupations from $63$ to less than $36$ with the number of eligible occupations varying per model. 
\subsection{Standard Error}
\label{appendix:standard-error}
\Cref{tab:standard-error} contains the standard error calculations for each occupation and model pair corresponding to the percent of biographies and personas associated with women. Standard errors were computed assuming a Bernoulli distribution, where each biography or persona is coded as either referencing a woman or not. All estimated standard errors are less than $0.003$. 

\begingroup
\captionsetup{width=0.95\textwidth}
\captionof{table}{The standard error corresponding with the percent of biographies and personas associated with women for each occupation and model pair. \textit{SE} refers to Standard Error.}
\label{tab:standard-error}
\begin{longtable}{@{}p{0.1305\textwidth}@{}p{0.29\textwidth}p{0.29\textwidth}@{}p{0.29\textwidth}@{}}

\toprule
Occupation & GPT-4o-mini SE & GPT-3.5 SE & LLama-3.1 SE \\
\midrule
\endfirsthead

\toprule
Occupation & GPT-4o-mini SE & GPT-3.5 SE & LLama-3.1 SE \\
\midrule
\endhead

\midrule
\multicolumn{2}{r}{\textit{Continued on next page}} \\
\midrule
\endfoot

\bottomrule
\endlastfoot
technician & $0.001971111952258775$ & $0.0013619246343602937$ & $0.0015037641213512564$\\
accountant & $0.0022439757386557243$ & $0.001742962092821394$ & $0.001363001730398828$\\
supervisor & $0.0021572660724440166$ & $0.0020915718840387766$ & $0.0015967558187937373$\\
engineer & $0.0023744118176713467$ & $0.0024504950741379553$ & $0.0024857510934416034$\\
worker & $0.0022998174118546467$ & $0.0022215433382394$ & $0.002024173236942328$\\
educator & $0.0006265810801888759$ & $0.0006092874163963133$ & $0.0$\\
clerk & $0.0021086369598665943$ & $0.002080190258875654$ & $0.0009234438756318972$\\
counselor & $0.0005954607326011406$ & $0.0006123253355583795$ & $0.0$\\
inspector & $0.0022503238108961144$ & $0.002350823094117581$ & $0.001946359828735317$\\
mechanic & $0.0$ & $0.00035355339059327376$ & $0.0$\\
manager & $0.0022217921840523088$ & $0.0017022920972453775$ & $0.0020208338955757478$\\
therapist & $0.0$ & $0.0$ & $0.0$\\
administrator & $0.001991931236462284$ & $0.0021165401420980274$ & $0.0003580574370197164$\\
salesperson & $0.0017151435005361834$ & $0.002292800729882237$ & $0.0019702240102655515$\\
receptionist & $0.0$ & $0.0$ & $0.0$\\
librarian & $0.0003758230140014144$ & $0.0$ & $0.0$\\
advisor & $0.0016593662957576616$ & $0.0019514019412319211$ & $0.001473071828578521$\\
pharmacist & $0.0007623498887196719$ & $0.0007034903951759158$ & $0.00036084391824351607$\\
janitor & $0.0012951324855997373$ & $0.0009234438756318972$ & $0.0010924593066487269$\\
psychologist & $0.0006092874163963133$ & $0.0006092874163963133$ & $0.0$\\
physician & $0.0008550764754654827$ & $0.0004987421363720583$ & $0.000505062920889691$\\
carpenter & $0.0003580574370197164$ & $0.0004987421363720583$ & $0.00035355339059327376$\\
nurse & $0.0$ & $0.0$ & $0.0$\\
investigator & $0.002387204841316108$ & $0.001955458237037049$ & $0.001793555626313547$\\
bartender & $0.002195023890535785$ & $0.0016034057716490862$ & $0.0025047541213593354$\\
specialist & $0.0011250718994483$ & $0.00036084391824351607$ & $0.0$\\
electrician & $0.0$ & $0.0009234438756318972$ & $0.00078450684658288$\\
officer & $0.0019083886158966137$ & $0.002420079320931029$ & $0.002498240586923276$\\
pathologist & $0.0$ & $0.000354440602504168$ & $0.0$\\
teacher & $0.0011279227649404297$ & $0.002005018828468342$ & $0.0$\\
lawyer & $0.0021397688440271184$ & $0.0018981110994045866$ & $0.001536613453189387$\\
planner & $0.0008614609845078961$ & $0.0008004201156468411$ & $0.0003904344047215152$\\
practitioner & $0.0014447592411016513$ & $0.0$ & $0.0005012418562445881$\\
plumber & $0.000372677996249965$ & $0.0$ & $0.0$\\
instructor & $0.0015642910679816596$ & $0.0009870203118662287$ & $0.000354440602504168$\\
surgeon & $0.001816331710021332$ & $0.0007034903951759158$ & $0.0011932248717822914$\\
veterinarian & $0.0006349440572278637$ & $0.0$ & $0.0$\\
paramedic & $0.0018424704325921719$ & $0.0023380357351719884$ & $0.0015037641213512564$\\
examiner & $0.0015324448103518095$ & $0.0009325664308941554$ & $0.0015530903093071255$\\
chemist & $0.0014223735573755572$ & $0.0$ & $0.0$\\
machinist & $0.0$ & $0.0$ & $0.0007034903951759158$\\
appraiser & $0.0021351724816583385$ & $0.002402570406790505$ & $0.0013264029237744874$\\
nutritionist & $0.0$ & $0.0$ & $0.0$\\
architect & $0.002473702750536783$ & $0.002013659592995872$ & $0.002202273283184153$\\
hairdresser & $0.0006680799186145912$ & $0.0007963512430110538$ & $0.0$\\
baker & $0.002038881180049122$ & $0.0004987421363720583$ & $0.00035355339059327376$\\
programmer & $0.001816763563307356$ & $0.0022558633133782043$ & $0.001989502926662312$\\
paralegal & $0.0006225345071547643$ & $0.0007141700498506129$ & $0.00036369648372665394$\\
hygienist & $0.0$ & $0.0$ & $0.0$\\
scientist & $0.0004987421363720583$ & $0.0$ & $0.0$\\
dispatcher & $0.0012106642658482037$ & $0.002491020189974577$ & $0.0006108007111619146$\\
cashier & $0.00045643546458763837$ & $0.0013161606988111314$ & $0.0$\\
auditor & $0.0023569065259169235$ & $0.002218901453797369$ & $0.0007883961087702325$\\
dietitian & $0.0$ & $0.0$ & $0.0$\\
painter & $0.0022594439967806262$ & $0.00035355339059327376$ & $0.0017392888282688139$\\
broker & $0.0025084590164469946$ & $0.002078255238947477$ & $0.0015287532952499377$\\
chef & $0.002451703494650308$ & $0.0021407742928352632$ & $0.002493710681860292$\\
doctor & $0.0012789364299916547$ & $0.0007017565899639197$ & $0.0$\\
firefighter & $0.0014516873457977586$ & $0.00235040617280516$ & $0.0008550764754654827$\\
secretary & $0.0$ & $0.0$ & $0.0$\\
software engineer & $0.0025033134540717967$ & $0.002090344009344693$ & $0.002420020243823993$\\
cook & $0.0018125717334141547$ & $0.002500951547250229$ & $0.0016857367952066444$\\
pilot & $0.0024161930441948636$ & $0.0016929931213284072$ & $0.0010924593066487269$\\
    \bottomrule
\end{longtable}

\subsection{Statistical Significance}
\label{appendix:stat-significance-GRBS}
\Cref{tab:stat-significance-GRBS} demonstrates the statistical significance of the Subset Representational Bias Scores by showing the p-values per model after running a t-test comparing SRB scores across occupation between women and men per model.
\begin{table}[H]
    \caption{Statistical significance of the Subset Representational Bias Scores per model.}
    \label{tab:stat-significance-GRBS}
    \centering
    \begin{tabular}{lll}
        \toprule
         Model & Welch's t-statistic & p-value\\
         \midrule
gpt-4o-mini-2024-07-18 & $-13.97$ & $4.028972846741477\mathrm{e}{-18}$\\
gpt-3.5-turbo-0125 & $-11.79$ & $1.4410110706776934\mathrm{e}{-16}$\\
LLama-3.1-70b & $-14.28$ & $2.2663766108258133\mathrm{e}{-18}$\\
        \bottomrule
    \end{tabular}
\end{table}

\Cref{tab:stat-significance-by-gender-btwn-gpt-3.5-and-gpt-4o-mini} demonstrates the statistical significance of the difference between the Subset Representational Bias Scores from GPT-3.5 to GPT-4o-mini. Here we run a t-test comparing the SRBS scores per gender from GPT-3.5 and GPT-4o-mini, and find that the p-values are less than $0.05$, indicating that the difference in scores is statistically significant.
\begin{table}[H]
    \caption{Statistical significance of the difference in Subset Representational Bias Scores per gender.}
    \label{tab:stat-significance-by-gender-btwn-gpt-3.5-and-gpt-4o-mini}
    \centering
    \begin{tabular}{lll}
        \toprule
        Gender & Welch's t-statistic & p-value\\
        \midrule
        F & $2.20$ & $0.03363462149929205$\\
        M & $-2.16$ & $0.036662054377888435$\\
        \bottomrule
    \end{tabular}
\end{table}

To compare the similarity of statistically significant words between associated men and women, we utilize the  methodology described in \Cref{sec:method-ssm+srbs} Chamfer Distance, as we cannot directly compare generations associated with men and women. Thus, we also generate $100$ personas per occupation, gender, and prompt, using the prompts where gender is specified to serve as our basis for comparison. The statistically significant words for specified gender are identified using the Calibrated Marked Words method per occupation and gender. Prior to using the Chamfer Distance, we remove pronouns from the statistically significant words, as differences in pronouns are expected. Our candidate sets are the word embeddings for the statistically significant words for associated men ($S_{AM}$) and women ($S_{AF}$), and our target sets are the word embeddings for the statistically significant words for specified men ($S_M$) and women ($S_F$). We use Word2Vec \cite{mikolov2013efficient} for our word embeddings.

We then utilize the Subset Representational Bias Scores to understand if there is a statistically significant difference in how associated men and associated women are described. We compare the $\Delta(S_{AF}\| S_F,S_M)$ for associated women and the $\Delta(S_{AM}\| S_F,S_M)$ for associated men which is between $-2$ and $2$. If $\Delta(S_{AF}\| S_F,S_M)$ is equivalent to $\Delta(S_{AM}\| S_F,S_M)$, this implies that there is no gendered difference in the statistically significant words for associated men and women. We find that the differences between $\Delta(S_{AF}\| S_F,S_M)$ and $\Delta(S_{AM}\| S_F,S_M)$ are statistically significant, as we compute the p-scores per model between the average Subset Representational Bias Score for each occupation between associated men and women. Each p-score was less than $0.05$, and the exact p-scores are provided in \Cref{tab:stat-significance-GRBS} in \Cref{appendix:stat-significance-GRBS}.

\section{Analysis}
\subsection{Who is represented}
\label{appendix:analysis-who}
As our results show in \Cref{sec:analysis-who}, GPT-3.5, GPT-4o-mini, and Llama-3.1 are more likely to generate biographies of women than with men, and this extends even to male-dominated occupations, where the majority are still primarily associated with women. On average, across occupations, the percentage of women exceeds that of men, and this trend seems to be more prominent for more recent models as shown in \Cref{fig:occupation_representation}. For instance, among male-dominated occupations, GPT-3.5 was much more likely to depict a small percentage of women, whereas GPT-4o-mini was more likely to depict majority women. Interestingly, the increase in female representation is pronounced across both male- and female-dominated occupations. However, this shift is not observed in traditionally male-dominated blue-collar occupations, such as technician, plumber, janitor, and carpenter, where female representation remains largely unchanged. While there are slight variations in gender association percentages based on the model and prompt used, the overall trend of increased female representation persists across all prompts, models, and occupations tested. \Cref{fig:prompt-specific-graphs} demonstrates the gender distribution breakdown based on prompt, demonstrating that despite slight differences in the distribution between prompts, the trend of increased representation of women holds on average across all prompts, models, and occupations tested. In \Cref{fig:occupation_representation} female representation per occupation is averaged across the two prompts used to generate the generations and female- and male-dominated occupations are determined by representation from the BLS.

\begin{figure}[h!]
    \centering
    \begin{subfigure}[t]{\textwidth}
        \centering
        \includegraphics[width=0.8\textwidth]{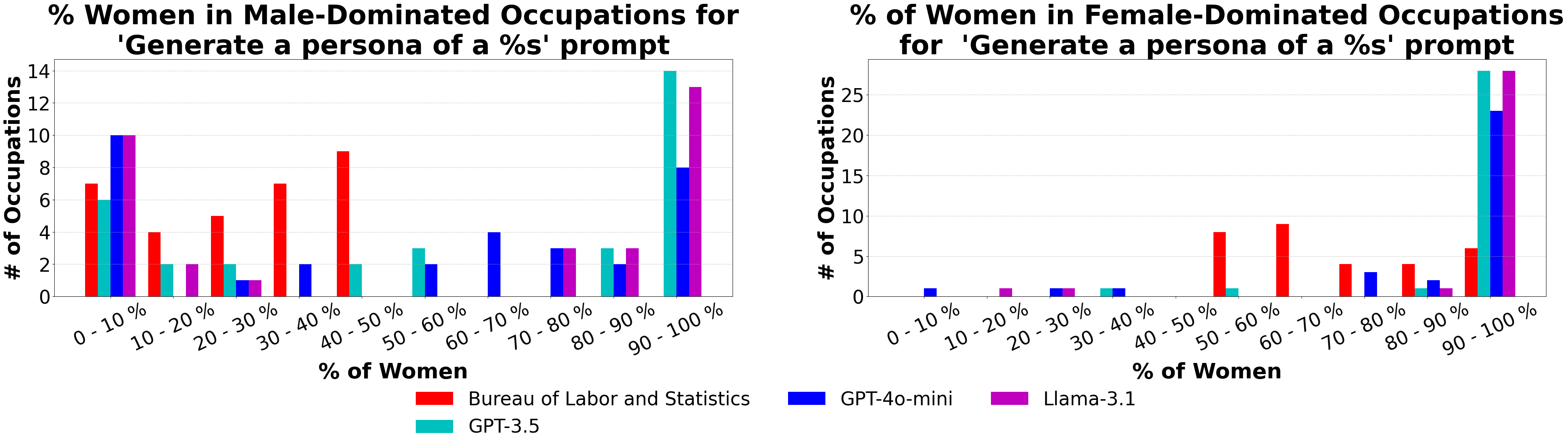}
        \caption{}
        \label{fig:persona}
    \end{subfigure}
    \vspace{1em} 

    \begin{subfigure}[t]{\textwidth}
        \centering
        \includegraphics[width=0.8\textwidth]{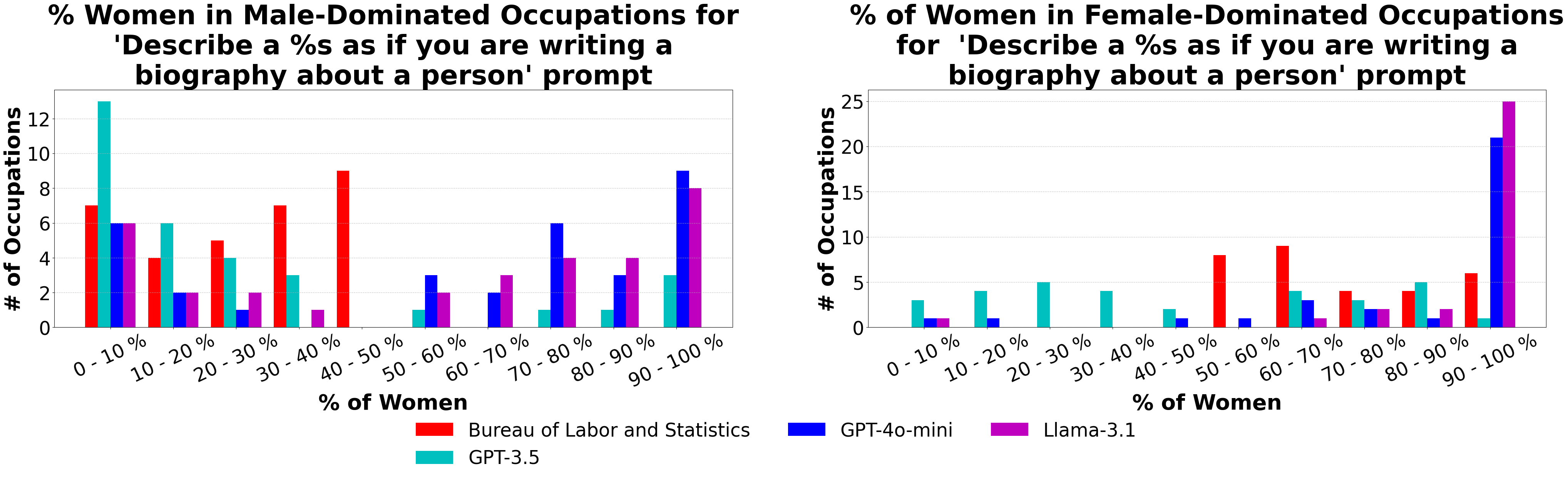}
        \caption{}
        \label{fig:biography}
    \end{subfigure}

    \caption{\% of women per occupation based on prompt and model in comparison to the Bureau of Labor and Statistics}
    \label{fig:prompt-specific-graphs}
\end{figure}
\subsubsection{Non-binary Representation}
\label{appendix:nb-rep}
\Cref{fig:nb-rep} demonstrates non-binary representation across occupations and models.
\begin{figure}
    \centering
    \includegraphics[width=0.5\linewidth]{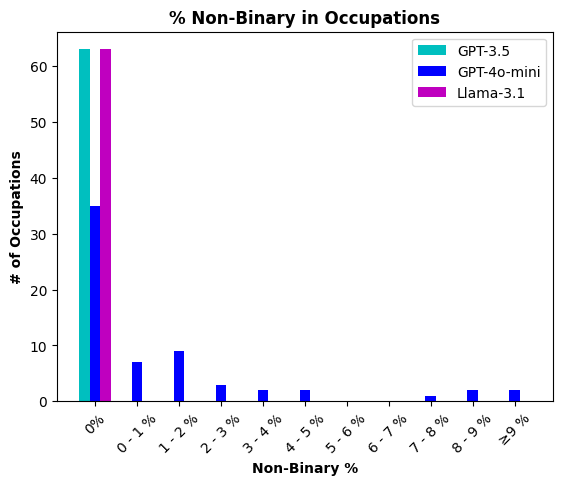}
    \caption{Percent of generations associated with non-binary per occupation based on model.}
    \label{fig:nb-rep}
\end{figure}
\subsection{How are people represented?}
A positive Subset Representational Bias Score is associated with men because $CH(S_{AM},S_{F})$ would be closer to $2$, as statistically significant words for associated men and specified women are not very similar. $CH(S_{AM},S_{F})$ would be closer to $0$, as statistically significant words for associated men and specified men would be similar. As $\Delta(S_{AM}\|S_{F},s_{M})=CH(S_{AM},S_F)-CH(S_{AM},S_M)$, a Subset Representational Bias Score that is positive indicates that associated men are more similar to specified men than women. A negative Subset Representational Bias Score for associated women indicates that associated women are more similar to specified women than men. 
\Cref{fig:subset-rep-bias-score} indicates that the statistically significant words for men and women differ as the Subset Representational Bias Scores across occupations and models for women are consistently negative, while the corresponding scores for men are higher than the scores for women.

\subsection{What are the implications of how people are represented?}
\label{appendix:implication}
In this section, we will first describe our clustering methodology in \Cref{appendix:clustering} and then provide a more detailed analysis of our findings in \Cref{appendix:what-further-analysis}.
\subsubsection{Clustering}
\label{appendix:clustering}
We use the K-means++ implementation in sklearn to cluster the statistically significant words (including pronouns identified as statistically significant) across model, occupation, and associated gender. To determine the optimal number of clusters to use, we use the Silhouette Score proposed by \citet{rousseeuw1987silhouettes}, as limitations with the Elbow Method for identifying the optimal number of clusters have been noted \cite{schubert2023stop}. The Silhouette Score measures how well an instance fits into a cluster. The score is between $-1$ and $1$, with a score of $1$ indicating that an instance is well defined for the cluster it is assigned, whereas a score of $-1$ indicates an instance was assigned the wrong cluster.

We plotted the Silhouette Statistic as shown in \Cref{fig:silhouette-graph} and determined that $1500$ clusters is the optimal number since it is the value of $k$ that has the largest gap between subsequent values of $k$ for our data. The word embeddings used for the identified statistically significant words are the M3-Embeddings \cite{chen2024m3}. Prior to running K-means++, we removed all names, one-letter words, two-letter words excluding `dr' and `md,' and non-English words.

\begin{figure}
    \centering
    \includegraphics[width=0.5\linewidth]{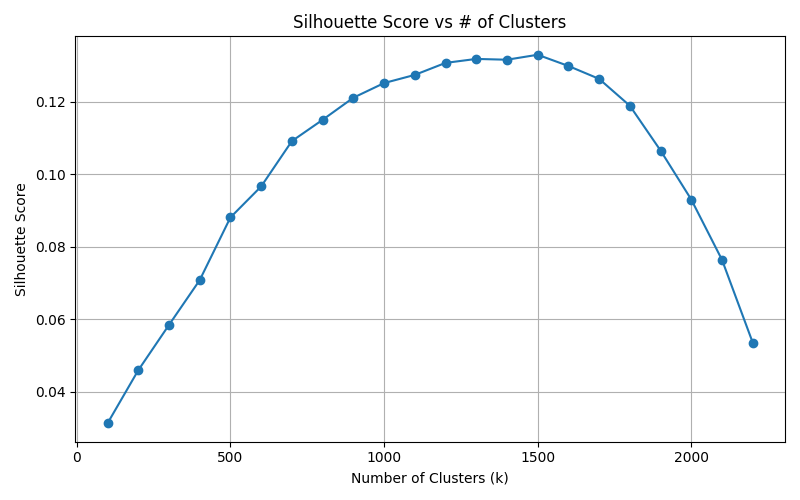}
    \caption{Silhouette score depending on the number of clusters.}
    \label{fig:silhouette-graph}
\end{figure}

After running K-means++ with $1500$ clusters, we identified patterns in the clusters that were at least $50$\% more prominent for one gender and appeared in at least three occupation-gender pairs per model of which there were $86$ clusters. The categories formed from these clusters are used in our analysis in \Cref{sec:analysis-what} and the corresponding words in each of these categories are shown in \Cref{fig:clusters}.

\subsubsection{Further Analysis}
\label{appendix:what-further-analysis}
Stereotypes as a representational harm have been outlined by numerous representational harm taxonomies \cite{chien2024beyond,katzman2023taxonomizing,shelby2023sociotechnical}. Gender stereotypes are prevalent across various contexts and can have harmful effects, whether the stereotype is perceived as positive or negative \cite{burkley2009positive,kay2013insidious}.
Stereotype-related words identified in our analysis focus primarily on the continuation of the association of empathy with women which is well documented in the literature \cite{christov2014empathy,loffler2023women}. 
Researchers note how discussion of excellence and academia support the meritocracy myth—the notion that success stems primarily from individual effort \cite{alvarado2010dispelling,blair2022misconceiving,lawton2000meritocracy,liu2011unraveling,razack2020beyond,rhode1996myths,staniscuaski2023science}. \citet{adamson2019female} and \citet{byrne2019role} discuss how media representation of ``inspirational" women emphasize that their achievements result primarily from individual effort, ignoring broader structural and systemic factors. Such narratives contribute to the perpetuation of systemic inequalities by obscuring the societal and institutional barriers that many face. 

\Cref{fig:clusters} illustrates that words such as ``inspire," ``inspires," and ``inspiring" are more frequently associated with women, and the manner in which these words are used in the female personas echos media discussion of inspirational women which emphasizes that their achievement, resulting from individual effort ignoring broader structural and systemic factors \cite{adamson2019female,byrne2019role}. 
\Cref{fig:clusters} shows a significantly higher prevalence of words related to advocacy and diversity such as ``advocate," ``diversity," and ``multicultural" for women compared to men across all occupations and models. 
Women and other underrepresented groups often feel or are pressured to represent their communities, participate in diversity initiatives, and mentor junior colleagues or students from underrepresented backgrounds \cite{altan2024care,ajayi2021prioritizing,docka2021twice,guarino2017faculty,malisch2020wake,misra2011ivory,porter2018burdens,pyke2015faculty,reid2021retaining,smith2023deciding,toole2019standpoint,zambrana2017blatant}.
The overrepresentation of advocacy-related words associated with women reinforces the expectation that women bear greater responsibility for advancing diversity and inclusion than men. This places the burden of addressing systemic inequities on women, rather than holding institutions and organizations accountable for meaningful change \cite{pyke2015faculty}. Achieving equity in the prevalence of advocacy-related language for both men and women would signal that advocacy and mentorship are collective responsibilities, not burdens to be disproportionately shouldered by marginalized groups. Recommendations for improving diversity, equity, and inclusion in workplaces and universities emphasize the importance of involving stakeholders from all groups and levels of the organization \cite{ueda2024emotional}.

\section{Calibrated Marked Words}
\label{sec:appendix-calibrated-marked-words}
The words displayed in \Cref{tab:calibrated-marked-words-software-engineer}
are the statistically significant words for software engineer identified using our Calibrated Marked Words method by model and gender. A full list of the words identified as statistically significant using the Calibrated Marked words method per model, occupation, and gender is available on \href{https://github.com/jennm/more-of-the-same}{Github} with the released code and results.

\begingroup
\captionsetup{width=0.95\textwidth}
\captionof{table}{The words displayed here are the statistically significant words for software engineer identified using the Calibrated Marked Words by model and gender. \textbf{Bolded words} are statistically significant for both the specified and associated generations (i.e., statistically significant for both associated and specified women), \underline{underlined words} are statistically significant for the associated gender and a different specified gender (i.e., statistically significant for associated women and specified men or vice versa), \gray{gray words} are only statistically significant for the specified gender, and all other words are only statistically significant for the associated gender.
}
\label{tab:calibrated-marked-words-software-engineer}
\begin{longtable}{@{}p{0.06\textwidth}p{0.06\textwidth}@{}p{0.9\textwidth}@{}}
\toprule
\textbf{Model} & \textbf{Gender} & \textbf{Generated Content }\\
\midrule
\endfirsthead

\toprule
\textbf{Model} & \textbf{Gender} & \textbf{Generated Content }\\
\midrule
\endhead

\midrule
\multicolumn{2}{r}{\textit{Continued on next page}} \\
\midrule
\endfoot

\bottomrule
\endlastfoot

gpt-4o-mini & F & \textbf{her}, \textbf{she}, \textbf{women}, \textbf{herself}, \textbf{diversity}, \textbf{stem}, \textbf{advocacy}, \textbf{careers}, \textbf{female}, \textbf{mit}, \textbf{minorities}, \textbf{shes}, \textbf{yoga}, \textbf{ava}, \textbf{young}, healthcare, \textbf{maledominated}, \textbf{unwavering}, \textbf{barriers}, \textbf{resilience}, \textbf{tech}, universitys, \textbf{technology}, alice, \textbf{countless}, \textbf{womenintech}, journey, stakeholders, \textbf{advocate}, \textbf{achievements}, \textbf{pursue}, \textbf{mental}, \textbf{inspired}, \textbf{empowering}, \textbf{forbes}, \textbf{syndrome}, \textbf{efforts}, ambitions, \textbf{undergraduate}, competitions, \textbf{imposter}, accomplishments, \textbf{numerous}, \textbf{inclusivity}, doe, personalized, initiatives, fathers, nontechnical, \textbf{inclusion}, \textbf{underrepresented}, father, \gray{dr}, \gray{girls}, \gray{inclusive}, \gray{field}, \gray{award}, \gray{pioneering}, \gray{phd}, \gray{organization}, \gray{promoting}, \gray{nonprofit}, \gray{advocating}, \gray{representation}, \gray{organizations}, \gray{stanford}, \gray{soughtafter}, \gray{initiative}, \gray{inspire}, \gray{empathetic}, \gray{pursuing}, \gray{california}, \gray{mentorship}, \gray{francisco}, \gray{san}, \gray{diverse}, \gray{codeher}, \gray{completing}, \gray{research}, \gray{processing}, \gray{workplace}, \gray{prestigious}, \gray{woman}, \gray{workshops}, \gray{multicultural}, \gray{aimed}, \gray{speaker}, \gray{communities}, \gray{empower}, \gray{practicing}, \gray{dissertation}, \gray{predominantly}, \gray{awards}, \gray{supportive}, \gray{studies}, \gray{encouraged}, \gray{equitable}, \gray{programs}, \gray{founded}, \gray{passionate}, \gray{accessibility}, \gray{journals}, \gray{faces}, \gray{scholarships}, \gray{instilled}, \gray{parents}, \gray{innovator}, \gray{trailblazing}, \gray{underserved}, \gray{painting}, \gray{anaya}, \gray{generations}, \gray{everyone}, \gray{berkeley}, \gray{focused}, \gray{determination}, \gray{immigrant}, \gray{conferences}, \gray{navigating}, \gray{break}, \gray{support}, \gray{equality}, \gray{resilient}, \gray{enter}, \gray{fostering}, \gray{proving}, \gray{doctoral}, \gray{volunteering}, \gray{model}, \gray{empowerment}, \gray{im}, \gray{bias}, \gray{aidriven}, \gray{gap}, \gray{promote}, \gray{contributions}, \gray{focuses}, \gray{networking}, \gray{biases}, \gray{womens}, \gray{empowered}, \gray{cum}, \gray{laude}, \gray{educators}, \gray{algorithms}, \gray{has}, \gray{resources}, \gray{publications}, \gray{summa}, \gray{featured}, \gray{recognized}, \gray{received}, \gray{institute}, \gray{collaborates}, \gray{confidence}, \gray{continue}, \gray{industry}, \gray{equity}, \gray{stereotypes}, \gray{academic}, \gray{panels}, \gray{traditionally}, \gray{extends}, \gray{for}, \gray{innovators}, \gray{perseverance}, \gray{rescue}, \gray{faced}, \gray{recognition}, \gray{massachusetts}, \gray{prowess}, \gray{others}, \gray{hiring}, \gray{luna}, \gray{health}, \gray{earning}, \gray{advancing}, \gray{thousands}, \gray{actively}, \gray{language}, \gray{tensorflow}, \gray{pave}, \gray{doctorate}, \gray{demonstrating}, \gray{immigrants}, \gray{accolades}, \gray{recipes}, \gray{future}, \gray{vocal}, \gray{nonprofits}, \gray{leadership} \\
& M & \textbf{his}, \textbf{he}, \textbf{him}, \textbf{male}, \textbf{himself}, \textbf{video}, \textbf{collaboration}, \textbf{gaming}, \textbf{architecture}, \textbf{games}, \textbf{indie}, \textbf{burgeoning}, highquality, brainstorming, \textbf{online}, nurturing, flagship, \textbf{gamer}, admired, \textbf{innovatech}, nathaniel, backgrounds, \textbf{fitness}, teammates, intricate, lucas, implementing, \textbf{playing}, embraced, forums, platforms, \textbf{tackling}, \textbf{adaptable}, gender, \textbf{kubernetes}, emphasizes, aidan, motivated, launching, learn, \textbf{software}, jon, \textbf{demographics}, willingness, realms, johnny, \textbf{hes}, multiple, \gray{code}, \gray{developers}, \gray{programming}, \gray{weekends}, \gray{projects}, \gray{development}, \gray{austin}, \gray{texas}, \gray{github}, \gray{knowledge}, \gray{languages}, \gray{businesses}, \gray{opensource}, \gray{junior}, \gray{friends}, \gray{solutions}, \gray{fastpaced}, \gray{keen}, \gray{values}, \gray{vision}, \gray{beginnings}, \gray{trends}, \gray{computers}, \gray{often}, \gray{complex}, \gray{technologies}, \gray{bustling}, \gray{efficient}, \gray{jacob}, \gray{enjoys}, \gray{deadlines}, \gray{colleagues}, \gray{struggles}, \gray{detailoriented}, \gray{analytical}, \gray{underprivileged}, \gray{lifestyle}, \gray{architect}, \gray{collaborative}, \gray{suburban}, \gray{multiplayer}, \gray{honed}, \gray{small}, \gray{boundaries}, \gray{venture}, \gray{management}, \gray{startup}, \gray{knack}, \gray{practices}, \gray{competitive}, \gray{avid}, \gray{spends}, \gray{contributes}, \gray{streamlined}, \gray{outdoor}, \gray{cycling}, \gray{communication}, \gray{youth}, \gray{graphic}, \gray{giving}, \gray{spent}, \gray{repositories}, \gray{weekend}, \gray{philosophy}, \gray{attracting}, \gray{aspirations}, \gray{apart}, \gray{peers}, \gray{frameworks}, \gray{streamline}, \gray{inspiration}, \gray{clean}, \gray{early}, \gray{exploring}, \gray{finding}, \gray{quality}, \gray{ability}, \gray{everevolving}, \gray{mobile}, \gray{manageable}, \gray{volunteered}, \gray{finds}, \gray{legacy}, \gray{clients}, \gray{immersed}, \gray{reviews}, \gray{solace}, \gray{problems}, \gray{collaborating}, \gray{designer}, \gray{interned}, \gray{healthy}, \gray{courses}, \gray{exhibited}, \gray{recharge}, \gray{tools}, \gray{graduating}, \gray{enthusiast}, \gray{stay}, \gray{ohio}, \gray{best}, \gray{downtime}, \gray{city}, \gray{pays}, \gray{style}, \gray{gadgets}, \gray{regularly}, \gray{feedback}, \gray{hours}, \gray{town}, \gray{allowed}, \gray{web}, \gray{propelled}, \gray{personal}, \gray{respected}, \gray{maintainable}, \gray{blogs}, \gray{updated}, \gray{years}, \gray{takes}, \gray{beauty}, \gray{prominence}, \gray{methodical}, \gray{developer}, \gray{biking}, \gray{databases}, \gray{marked}, \gray{collaborated}, \gray{java}, \gray{entrepreneurial}, \gray{likes}, \gray{push}, \gray{believing}, \gray{penchant}, \gray{vibrant}, \gray{podcasts}, \gray{flourished}, \gray{expertise}, \gray{learner}, \gray{prefers}, \gray{enthusiasts}, \gray{team}, \gray{nate}, \gray{adventures}, \gray{contributing}, \gray{latest}, \gray{continuous}, \gray{with}, \gray{larger}, \gray{player}, \gray{thinker}, \gray{superiors}, \gray{likeminded}, \gray{optimizing}, \gray{startups}, \gray{apps}, \gray{processes}, \gray{mongodb}, \gray{agile}, \gray{simple}, \gray{outdoors}, \gray{reputation}, \gray{activities}, \gray{designing}\\
gpt-3.5 & F & \textbf{she}, \textbf{her}, reading, \textbf{alice}, \textbf{herself}, \textbf{diversity}, \textbf{inclusion}, \textbf{careers}, practicing, hiking, math, nontechnical, creative, \textbf{excelled}, \textbf{yoga}, \textbf{inspire}, engineering, \underline{expand}, \textbf{pursue}, \gray{women}, \gray{stem}, \gray{advocate}, \gray{female}, \gray{maledominated}, \gray{actively}, \gray{determination}, \gray{facing}, \gray{barriers}, \gray{pursuing}, \gray{promoting}, \gray{interested}, \gray{gender}, \gray{young}, \gray{woman}, \gray{trailblazing}, \gray{biography}, \gray{tech}, \gray{industry}, \gray{anna}, \gray{model}, \gray{perseverance}, \gray{breaking}, \gray{stereotypes}, \gray{advocating}, \gray{discovered}, \gray{obstacles}, \gray{empower}, \gray{town}, \gray{proving}, \gray{equality}, \gray{other}, \gray{inspiring}, \gray{girls}, \gray{break}, \gray{everywhere}, \gray{discrimination}, \gray{serves}, \gray{confident}, \gray{testament}, \gray{unwavering}, \gray{way}, \gray{shattered}, \gray{workplace}, \gray{journey}, \gray{involved}, \gray{underrepresented}, \gray{empowering}, \gray{delve}, \gray{growing}, \gray{inspiration}, \gray{organizations}, \gray{intelligence}, \gray{aspiring}, \gray{generations}, \gray{promote}, \gray{encouraged}, \gray{landed}, \gray{resilience}, \gray{mentorship}, \gray{inspired}, \gray{passions}, \gray{paved}, \gray{supportive}, \gray{programs}, \gray{initiatives}, \gray{pursued}, \gray{support}, \gray{platform}, \gray{pioneering}, \gray{fierce}, \gray{encouraging}, \gray{paving}, \gray{along}, \gray{determined}, \gray{small}, \gray{footsteps}, \gray{field}, \gray{role}, \gray{propelled}, \gray{biases}, \gray{up}, \gray{countless}, \gray{anything}, \gray{vocal}, \gray{captivating}, \gray{tenacity}, \gray{adversity}, \gray{mathematics} \\
& M & \textbf{he}, \textbf{his}, \textbf{playing}, \textbf{video}, \textbf{games}, male, \underline{gender}, strives, \gray{him}, \gray{enjoys}, \gray{expertise}, \gray{programming}, \gray{technologies}, \gray{avid}, \gray{highly}, \gray{himself}, \gray{technical}, \gray{staying}, \gray{knowledge}, \gray{craft}, \gray{ethic}, \gray{spare}, \gray{thinker}, \gray{latest}, \gray{projects}, \gray{software}, \gray{accomplished}, \gray{developers}, \gray{logical}, \gray{known}, \gray{achieve}, \gray{solutions}, \gray{dedicated}, \gray{silicon}, \gray{constantly}, \gray{exploring}, \gray{trends}, \gray{improve}, \gray{collaborative}, \gray{sharing}, \gray{skills}, \gray{detailoriented}, \gray{willing}, \gray{push}, \gray{skilled}, \gray{companies}, \gray{hackathons}, \gray{complex}, \gray{overall}, \gray{google}, \gray{analytical}, \gray{with}, \gray{gamer}, \gray{valley}, \gray{opensource}, \gray{attending}, \gray{always}, \gray{junior}, \gray{boundaries}, \gray{everevolving}, \gray{biggest}, \gray{husband}, \gray{development}, \gray{ability}, \gray{online}, \gray{contributing}, \gray{spending}, \gray{seasoned}, \gray{languages}, \gray{efficient}, \gray{honing}, \gray{balance}, \gray{striving}, \gray{camping}, \gray{seeking}, \gray{working}, \gray{new}, \gray{asset}, \gray{deadlines}, \gray{competitions}, \gray{reputation}, \gray{looking}, \gray{obtaining}, \gray{worklife}, \gray{share}, \gray{finding}, \gray{admire}, \gray{uptodate}, \gray{excellence}, \gray{hobbies}, \gray{ensure}, \gray{team}, \gray{methodical}, \gray{lasting}, \gray{project}, \gray{outdoor}, \gray{additionally}, \gray{exposed}, \gray{soughtafter}, \gray{eric}, \gray{mastering}, \gray{engineer}, \gray{father}, \gray{expand}, \gray{digital}, \gray{enhance}, \gray{work}, \gray{frameworks}, \gray{raised}, \gray{collaborates}, \gray{clients}\\
llama-3.1 & F & \textbf{her}, \textbf{she}, \textbf{herself}, \textbf{women}, \textbf{diversity}, \textbf{inclusion}, \textbf{alexandra}, \textbf{trailblazing}, \textbf{empowering}, \textbf{underrepresented}, \textbf{yoga}, talent, \textbf{tech}, fulltime, \textbf{nonprofit}, internship, \textbf{careers}, \textbf{practicing}, \textbf{stem}, companys, \textbf{groups}, \textbf{promote}, \textbf{lexi}, collaborative, \textbf{stanford}, coveted, spot, technologists, intensified, \textbf{prestigious}, \gray{pursue}, \gray{dr}, \gray{advocate}, \gray{technology}, \gray{industry}, \gray{research}, \gray{processing}, \gray{award}, \gray{phd}, \gray{language}, \gray{ava}, \gray{indian}, \gray{maledominated}, \gray{india}, \gray{promoting}, \gray{encouraged}, \gray{aipowered}, \gray{parents}, \gray{nalini}, \gray{ruku}, \gray{akira}, \gray{girls}, \gray{institute}, \gray{inspire}, \gray{inclusive}, \gray{woman}, \gray{natural}, \gray{indianamerican}, \gray{rohini}, \gray{model}, \gray{nationality}, \gray{shes}, \gray{intelligence}, \gray{mentorship}, \gray{mellon}, \gray{focused}, \gray{assistant}, \gray{undergraduate}, \gray{carnegie}, \gray{mumbai}, \gray{organizations}, \gray{machine}, \gray{meditation}, \gray{organization}, \gray{selfdoubt}, \gray{artificial}, \gray{young}, \gray{support}, \gray{mathematics}, \gray{stereotypes}, \gray{mit}, \gray{tireless}, \gray{interaction}, \gray{determination}, \gray{empathetic}, \gray{advocacy}, \gray{forbes}, \gray{biases}, \gray{syndrome}, \gray{empower}, \gray{overcoming}, \gray{dissertation}, \gray{impact}, \gray{imposter}, \gray{thesis}, \gray{expressive}, \gray{female}, \gray{science}, \gray{equity}, \gray{initiatives}, \gray{scholarship}, \gray{for}, \gray{vocal}, \gray{academic}, \gray{barriers}, \gray{anita}, \gray{bias}, \gray{paved}, \gray{moved}, \gray{supportive}, \gray{confident}, \gray{cuisines}, \gray{nali}, \gray{massachusetts}, \gray{journals}, \gray{humancomputer}, \gray{womens}, \gray{immigrant}, \gray{learning}, \gray{borg}, \gray{education}, \gray{seoul}, \gray{recipes}, \gray{excelled}, \gray{computer}, \gray{united}, \gray{pioneering}, \gray{navigating}, \gray{workplace}, \gray{korea}, \gray{joined}, \gray{leader}, \gray{generations}, \gray{graduate}, \gray{honors}, \gray{virtual}, \gray{accessible}, \gray{empowerment}, \gray{studies}, \gray{university}, \gray{acm}, \gray{states}, \gray{jewelry}, \gray{foundations}, \gray{mexico}, \gray{pytorch}, \gray{math}, \gray{role}, \gray{field}, \gray{mentions}, \gray{alisha}, \gray{selfcare}, \gray{fluent}, \gray{social}, \gray{tensorflow}, \gray{earned}, \gray{frankly}, \gray{colorful}, \gray{national}, \gray{minorities}, \gray{interfaces}, \gray{challenges}, \gray{researchers}, \gray{top}, \gray{google}, \gray{volunteering}, \gray{cooking}, \gray{valued}, \gray{awarded}, \gray{founded}, \gray{advocating}, \gray{acclaimed}, \gray{equitable}, \gray{instilled}, \gray{positive}, \gray{warm}, \gray{balancing}, \gray{hopper}, \gray{assertive}, \gray{microsoft}, \gray{descent}, \gray{diverse}, \gray{leading}, \gray{programs}, \gray{aimed}, \gray{curly}, \gray{encouraging}, \gray{recognition}, \gray{received}, \gray{vision}, \gray{leela}, \gray{researcher}, \gray{toptier}, \gray{educators}, \gray{representation}, \gray{alee}, \gray{anitaborg} \\
& M & \textbf{his}, \textbf{he}, \textbf{him}, \textbf{himself}, \textbf{code}, craftsman, ecommerce, gadgets, \textbf{embarked}, problemsolver, \textbf{disassembling}, prominent, \textbf{development}, jds, guiding, pursuit, \textbf{architect}, fortune, \textbf{programming}, \textbf{realm}, \textbf{languages}, \gray{software}, \gray{playing}, \gray{games}, \gray{short}, \gray{guitar}, \gray{beard}, \gray{jeans}, \gray{casual}, \gray{hes}, \gray{lean}, \gray{blue}, \gray{video}, \gray{introverted}, \gray{projects}, \gray{lbs}, \gray{complex}, \gray{build}, \gray{mobile}, \gray{enjoys}, \gray{latest}, \gray{avid}, \gray{height}, \gray{ability}, \gray{music}, \gray{c}, \gray{engineering}, \gray{landed}, \gray{honed}, \gray{computers}, \gray{strategy}, \gray{with}, \gray{spent}, \gray{firm}, \gray{weight}, \gray{django}, \gray{tshirts}, \gray{java}, \gray{enthusiast}, \gray{cm}, \gray{basketball}, \gray{humble}, \gray{sneakers}, \gray{silicon}, \gray{detailoriented}, \gray{hair}, \gray{scifi}, \gray{junior}, \gray{hoodies}, \gray{reassembling}, \gray{analytical}, \gray{collaborate}, \gray{buttondown}, \gray{occupation}, \gray{mongodb}, \gray{uptodate}, \gray{postgresql}, \gray{macos}, \gray{highly}, \gray{technologies}, \gray{app}, \gray{mysql}, \gray{grow}, \gray{facial}, \gray{linux}, \gray{knowledge}, \gray{opensource}, \gray{python}, \gray{attire}, \gray{devoted}, \gray{windows}, \gray{style}, \gray{peers}, \gray{trimmed}, \gray{kg}, \gray{crossfunctional}, \gray{skills}, \gray{gaming}, \gray{collaborating}, \gray{valley}, \gray{hours}, \gray{watching}, \gray{messy}, \gray{landscape}, \gray{advancements}, \gray{ryder}, \gray{meetups}, \gray{hobbies}, \gray{wellgroomed}, \gray{hiking}, \gray{insatiable}, \gray{stay}, \gray{affinity}, \gray{bit}, \gray{spring}, \gray{applications}, \gray{online}, \gray{workings}, \gray{solving}, \gray{systems}, \gray{shirts}, \gray{react}, \gray{problems}, \gray{neatly}, \gray{perfectionist}, \gray{remains}, \gray{expertise}, \gray{teams}, \gray{likeminded}, \gray{operating}, \gray{physical}, \gray{graduating}, \gray{immersed}, \gray{multiple}, \gray{tall}, \gray{spending}, \gray{developers}, \gray{humor}, \gray{cybersecurity}, \gray{detail}, \gray{skilled}, \gray{collaborated}, \gray{reputable}, \gray{dresses}, \gray{graduation}, \gray{forefront}, \gray{honing}, \gray{player}, \gray{colorado}, \gray{mountains}, \gray{scalable}, \gray{trends}, \gray{contributing}, \gray{solutions}, \gray{priorities}, \gray{francisco}, \gray{job}, \gray{outdoor}, \gray{range}, \gray{unwavering}, \gray{fascination}, \gray{successful}, \gray{apps}, \gray{pc}, \gray{techindustry}, \gray{various}, \gray{typically}, \gray{graphic}, \gray{tapping}, \gray{lucas}, \gray{movies}, \gray{reader}, \gray{revolutionized}, \gray{frameworks}, \gray{learn}, \gray{tinkering}, \gray{midsized}, \gray{deadlines}, \gray{intricacies}, \gray{engineer}, \gray{responsibilities}, \gray{reputation}, \gray{features}, \gray{participated}, \gray{chilly}, \gray{technological}, \gray{senior}, \gray{san}, \gray{outdoors}, \gray{dry}, \gray{craft}, \gray{intellij}, \gray{gamer}, \gray{introvert}, \gray{wife}, \gray{despite}, \gray{tshirt}, \gray{hone}, \gray{delegating}, \gray{accomplishments}, \gray{clients}, \gray{innovation}, \gray{demonstrated}\\
\bottomrule
\end{longtable}
\endgroup

\end{document}